\documentclass[sigconf]{acmart}
\usepackage{multirow}
\usepackage[table]{xcolor}
\usepackage{fontawesome}
\usepackage{graphicx}   % 插入图片必需
\usepackage{subcaption} % 子图功能（推荐）

\AtBeginDocument{%
  }

\copyrightyear{2026}
\acmYear{2026}
\setcopyright{cc}
\setcctype{by}
\acmConference[MM '26]{Proceedings of the 34th ACM International Conference on Multimedia}{November 10--14, 2026}{Rio de Janeiro, Brazil}
\acmBooktitle{Proceedings of the 34th ACM International Conference on Multimedia (MM '26), November 10--14, 2026, Rio de Janeiro, Brazil}
\acmDOI{10.1145/3767308.3834965}
\acmISBN{979-8-4007-2213-4/2026/11}
\begin{document}

%%
%% The "title" command has an optional parameter,
%% allowing the author to define a "short title" to be used in page headers.
\title{PixARG: Pixel-Asynchronous Retrieval-Augmented Diffusion for Fast and Realistic Image Generation}

%%
%% The "author" command and its associated commands are used to define
%% the authors and their affiliations.
%% Of note is the shared affiliation of the first two authors, and the
%% "authornote" and "authornotemark" commands
%% used to denote shared contribution to the research.
\settopmatter{authorsperrow=4}
\author{Zewei Xin}
\affiliation{
  \institution{Shanghai Jiao Tong University}
  \city{Shanghai}
  \country{China}}
\email{xin\_zewei\_12138@sjtu.edu.cn}
\author{Qinya Li}
\authornote{Qinya Li and Chaoyue Niu are corresponding authors.}
\correspondingauthor
\affiliation{
  \institution{Shanghai Jiao Tong University}
  \city{Shanghai}
  \country{China}}
\email{qinyali@sjtu.edu.cn}
\author{Chaoyue Niu}
\authornotemark[1]
\correspondingauthor
\affiliation{
  \institution{Shanghai Jiao Tong University}
  \city{Shanghai}
  \country{China}}
\email{rvince@sjtu.edu.cn}
\author{Fan Wu}
\affiliation{
  \institution{Shanghai Jiao Tong University}
  \city{Shanghai}
  \country{China}}
\email{fwu@cs.sjtu.edu.cn}

\begin{abstract}
  While diffusion models achieve impressive image generation capabilities, they suffer from high computational costs, implausible artifacts, and limited adaptability to dynamic or private data due to static, centralized training. To address these, we propose PixARG, a novel retrieval-augmented generative framework that integrates dynamic real-world image retrieval into the diffusion pipeline to simultaneously enhance efficiency and visual realism. Specifically, PixARG constructs a pixel-level affinity map between the user's text prompt and the retrieved image to guide a heterogeneous noise initialization. By modulating noise intensity across spatial regions, this mechanism preserves semantically relevant structural priors while perturbing less relevant areas, thereby significantly accelerating convergence. To accommodate this spatially non-uniform noise and ensure synchronized generation, we introduce a pixel-wise asynchronous denoising scheduler and a spatiotemporal denoiser conditioned on the affinity map. Furthermore, we derive an affinity-adaptive training objective that elegantly balances deterministic restoration with stochastic regeneration across pixels. Experiments on LAION and COYO datasets across offline, online, and user-specific settings show PixARG enhances realism and semantic alignment while drastically reducing diffusion steps.
\end{abstract}

%%
%% The code below is generated by the tool at http://dl.acm.org/ccs.cfm.
%% Please copy and paste the code instead of the example below.
%%


%% A "teaser" image appears between the author and affiliation
%% information and the body of the document, and typically spans the
%% page.
% \begin{teaserfigure}
%  \includegraphics[width=\textwidth]{sampleteaser}
%   \caption{Seattle Mariners at Spring Training, 2010.}
%   \Description{Enjoying the baseball game from the third-base
%   seats. Ichiro Suzuki preparing to bat.}
%   \label{fig:teaser}
% \end{teaserfigure}

% \received{20 February 2007}
% \received[revised]{12 March 2009}
% \received[accepted]{5 June 2009}

%%
%% This command processes the author and affiliation and title
%% information and builds the first part of the formatted document.
\maketitle

\section{Introduction}

The powerful generative capability of diffusion models~\cite{sd3,imagen3,FLUX.1Kontext,dalle3} enables users to create diverse high-quality images given text prompts that reflect their intended concepts.
However, diffusion models demand substantial computational resources, posing significant challenges for deployment on resource-constrained devices, especially consumer-grade hardware. The iterative denoising process and massive parameterization result in high latency and memory consumption. To mitigate this, recent research improves inference efficiency through techniques such as step reduction~\cite{yin2024one,luo2024one}, parameter compression~\cite{zhang2024laptop,li2024svdquant}, and computation reuse~\cite{shen2025lazydit,kahatapitiya2025adaptive}.
However, these acceleration approaches typically rely on constrained parameter budgets and simplified structures that often compromise the model expressivity and generation quality, leading to performance degradation and limited applicability in generative tasks.

\begin{figure}[!t]
    \centering
    \includegraphics[width=0.9\linewidth]{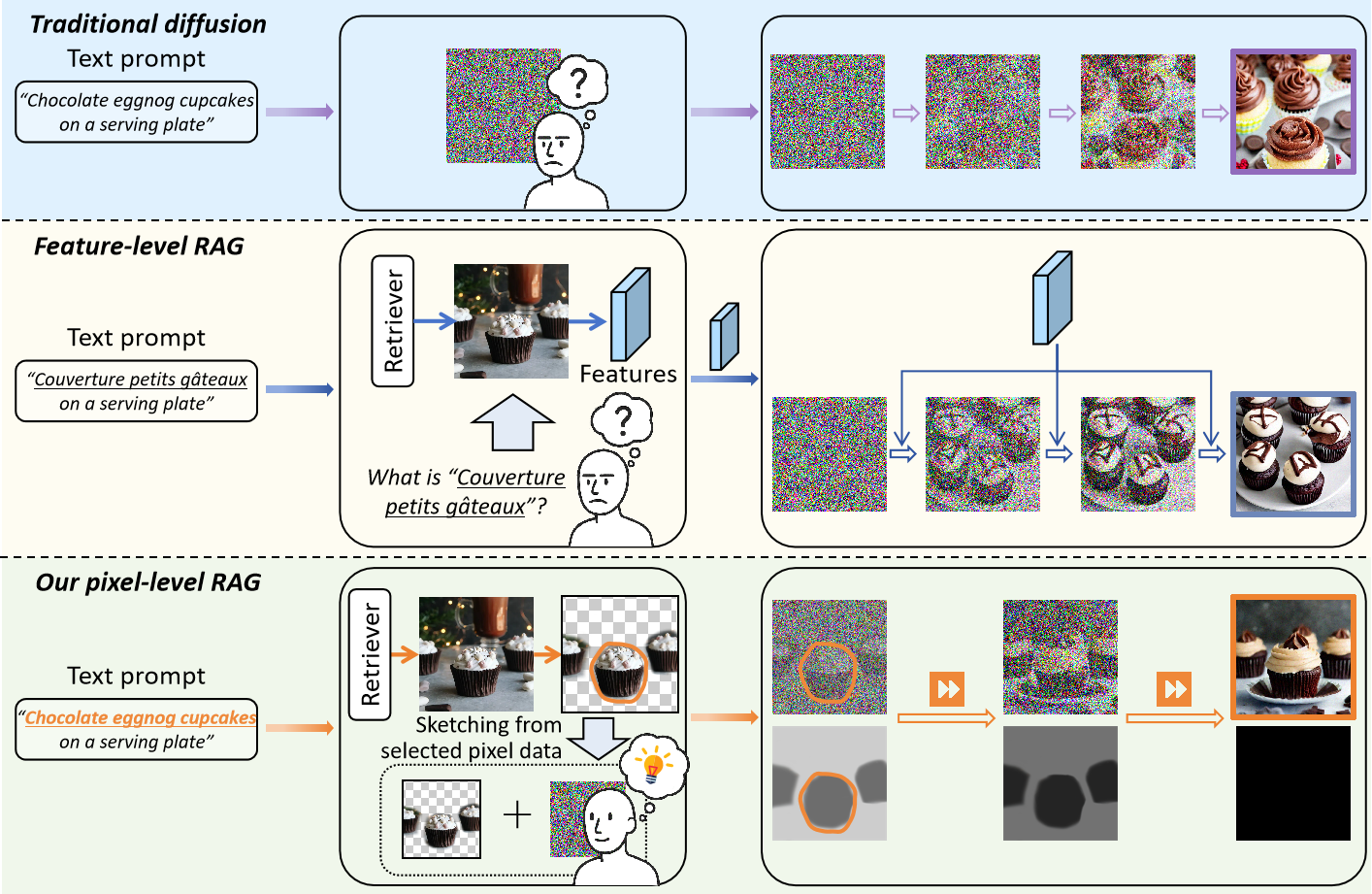}
    \Description{Comparison among diffusion methods. Traditional diffusion generates images from pure noise, and feature-level RAG diffusion incorporates retrieved image features into the denoising network to support new concepts. In contrast, our pixel-level RAG retrieves prompt-relevant pixel information from real images and injects it into the initial noise. This introduces realistic texture, structure and other low-level cues to enhance fidelity, while also accelerates generation through an information-enriched initialization that shortens the effective denoising trajectory.}
    \caption{Comparison among diffusion methods. Traditional diffusion generates images from pure noise, and feature-level RAG diffusion incorporates retrieved image features into the denoising network to support new concepts. In contrast, our pixel-level RAG retrieves prompt-relevant pixel information from real images and injects it into the initial noise. This introduces realistic texture, structure and other low-level cues to enhance fidelity, while also accelerates generation through an information-enriched initialization that shortens the effective denoising trajectory.}
    \label{fig:intro}
\end{figure}

Concurrently, Retrieval-Augmented Generation (RAG)~\cite{lewis2020retrieval,gao2023retrieval} has achieved remarkable success in large language models (LLMs) by coupling generation with external knowledge retrieval. By offloading the knowledge burden to external data sources, RAG enables lightweight models to achieve performance comparable to much larger models, thereby significantly improving efficiency~\cite{izacard2023atlas,huang2024less}. This decoupling of knowledge from static parameters also allows LLMs to integrate newly acquired information without the need for retraining, enhancing factual consistency and reducing hallucinations~\cite{kashmira2024tobugraph,guan2025deeprag}. Furthermore, RAG facilitates personalized, high-quality generation by leveraging local user data while keeping sensitive information on-device~\cite{wang2024mememo,wang2024crafting,salemi2026improving}. This effectively addresses a critical limitation of centralized training paradigms, where such private, dynamic data is fundamentally inaccessible. While recent studies have begun exploring the integration of RAG with text-to-image diffusion models to synthesize novel concepts and unseen compositions~\cite{blattmann2022retrieval,sheynin2022knn,chen2022re}, the potential of utilizing retrieved images to simultaneously improve the efficiency, realism and scalability of the generative process remains largely unexplored.

In this work, we propose a retrieval-driven generative framework that unifies accelerated inference with enhanced visual fidelity (Fig. \ref{fig:intro}). We argue that RAG is not merely an auxiliary data source, but a critical catalyst for diffusion efficiency, quality, and personalization. Specifically, our framework leverages retrieval in three key ways: (1) First, the retrieved images provide abundant low-level structural and textural priors, offering pixel-level constraints and initialization cues that allow the diffusion process to converge faster with reduced denoising uncertainty. (2) Second, these retrieved images act as real-world exemplars that inherently obey natural semantic and physical regularities; incorporating such priors effectively mitigates unrealistic or implausible artifacts in the generated results. (3) Finally, integrating RAG empowers the diffusion pipeline to exploit massive, continuously updated private data streams that remain inaccessible to centralized training, enabling highly tailored and user-specific generation. By reimagining RAG within the diffusion pipeline, we provide a promising pathway toward a highly efficient, realistic, and adaptive generative system that transcends the limitations of static, parameter-heavy models.

Technically, we propose pixel-asynchronous retrieval-augmented generation (PixARG). The key insight is to leverage external images to provide real-world information for improving generation realism, and to use their low-level visual features and structural information as initialization cues, thereby accelerating generation.
First, to precisely retrieve the necessary real-world information, we use the given text prompt to retrieve the most relevant images from a image database and construct a pixel-level text-visual affinity map to pinpoint the high-relevance regions between the retrieved image and the text description.
Then, based on the semantic alignment defined by this affinity map, PixARG injects spatially heterogeneous noise, which selectively preserves visual pixel information in semantically relevant regions while corrupts low-relevance areas. In this way, PixARG constructs a visual prior containing heterogeneous noise as the initial latent and begins the denoising process from it.
Subsequently, to ensure that regions with different initial noise intensities can converge consistently during the sampling process, we design a pixel-wise asynchronous denoising scheduler capable of handling this spatially non-uniform noise. Furthermore, we also introduce this affinity map into the conditioning of the noise prediction model and design a corresponding loss function, enabling the model to perceive and effectively handle this heterogeneous noise.

Our key contributions are as follows:

\begin{itemize}
    \item We propose Pixel-Asynchronous Retrieval-Augmented Generation (PixARG), the first framework that leverages pixel-level information from retrieved images to enable fast and realistic text-to-image generation.

    \item We construct an text-visual affinity map capturing pixel-wise semantic relevance to guide heterogeneous noise initialization by selectively preserving relevant regions while injecting noise elsewhere. The affinity map further conditions a noise predictor with a tailored loss, while a pixel-wise asynchronous denoising schedule handles spatial heterogeneity to achieve coherent and realistic generation.

    \item Our PixARG, built upon SD 1.4 backbone, achieves a 2.4x speedup on the LAION and COYO datasets while simultaneously improving realism metrics, with precision increasing by 9.12-11.94\% and plausibility by 1.04-3.52\%. 
    % Compared to SDXL Turbo, it uses only 33\% of the parameters, achieves a 0.39 gain in CLIP score, and runs 3.8x faster on SD1.4 backbone.
    Built on SD1.4, our method outperforms SDXL Turbo using only 33\% of the parameters, achieves a 0.39 gain in CLIP, and runs 3.8x faster.
\end{itemize}

\section{Related Work}

\subsection{Acceleration of Diffusion Models}

Some existing work accelerate inference by enabling diffusion model to complete generation in fewer steps through distillation.
InstaFlow~\cite{liu2023instaflow} adopts rectified flow distillation to align noise and image trajectories, yielding a one-step diffusion model.
DMD~\cite{yin2024one} further introduces distribution-matching distillation to ensure consistency between one-step generators and multi-step teachers.

Some researches focus on making diffusion models more lightweight. Quantization methods reduce computational and memory overhead by using lower-precision formats.
Q-Diffusion~\cite{li2023q} introduces a timestep-aware post-training quantization framework for 4-bit inference.
SVDQuant~\cite{li2024svdquant} combines singular value decomposition with 4-bit quantization for enhanced compression.
In addition, pruning removes redundant parameters to simplify architectures.
Hybrid SD~\cite{yan2024hybrid} employs structured pruning for edge deployment.
TinyFusion~\cite{fang2025tinyfusion} performs learnable depth pruning for diffusion transformers, while LAPTOP-Diff~\cite{zhang2024laptop} integrates one-shot layer pruning with normalized feature distillation to maintain quality.

From computation reuse, caching methods reuses or forecasts intermediate features without fine-tuning.
DeepCache~\cite{ma2024deepcache} reuses deep features across similar timesteps to eliminate redundant computation,
whereas TaylorSeers~\cite{liu2025reusing} predicts future features using Taylor expansion.
Similarly, ReDi~\cite{zhang2023redi} retrieves and reuses previously generated denoising trajectories to replace current ones.

While existing acceleration techniques improve efficiency, they are limited by the base model's capacity, frequently causing quality degradation.
As reported, TinyFusion~\cite{fang2025tinyfusion} observes an FID increase from 2.27 to 2.55, DMD~\cite{yin2024one} from 2.32 to 2.62, and TaylorSeers~\cite{liu2025reusing} from 2.32 to 2.65.

\subsection{RAG for Diffusion Models}

KNN-Diffusion~\cite{sheynin2022knn} introduces a retrieval-based diffusion framework that conditions the model jointly on the CLIP embedding of the prompt and its $k$ nearest retrieved images. This broadens the conditional distribution and enables cross-domain generalization simply by replacing the retrieval database.
Similarly, RDM~\cite{blattmann2022retrieval} conditions the model on retrieved CLIP image features and supports flexible combinations of textual and visual embeddings. It further demonstrates that the model can transfer to new domains by replacing the retrieval database. 
Re-Imagen~\cite{chen2022re} extends RAG to a multimodal knowledge base by retrieving the top-$k$ image-text pairs. After being encoded by the same downsampling path of the U-Net, the retrieved images are fused with the latent via cross-attention to inject visual priors, while the retrieved textual descriptions are used as joint conditioning with the query prompt during the upsampling stage. This interleaved strategy ensures joint text–image alignment, improving the fidelity of rare or unseen entities.

Existing retrieval-augmented methods primarily focus on broadening semantic coverage by using encoded image features, but rarely leverage retrieved images to jointly enhance realism and efficiency.
In contrast, PixARG incorporates real images as pixel-level priors into diffusion process, exploiting their authentic structural and pixel information to achieve realistic and efficient generation.

\section{Preliminaries: Latent Diffusion Models for Text-to-Image Generation}

Latent Diffusion Models (LDMs) perform the diffusion process in a compressed latent space.
A pre-trained autoencoder, composed of an encoder $\mathcal{E}$ and a decoder $\mathcal{D}$, compresses an image $v$ into a latent representation $x_0 = \mathcal{E}(v)$ and reconstructs it as $\hat{v} = \mathcal{D}(x_0)$.
All diffusion operations are performed in this latent domain.

In the forward process, the clean latent $x_0$ is gradually corrupted by the addition of Gaussian noise over $T$ timesteps, such that $x_T$ approximates an isotropic Gaussian distribution. 
Under the standard Gaussian assumption, the noisy latent at step $t$ is defined as:
\begin{align}
    x_t = a_t x_0 + b_t \epsilon_t,
    \label{eq:forward}
\end{align}
where $\epsilon_t \sim \mathcal{N}(0, \mathbf{I})$, and $a_t$ and $b_t$ are deterministic coefficients derived from a predefined noise schedule.

Leveraging this formulation, the model $\epsilon_\theta(x_t, t, c)$ is trained to predict the noise $\epsilon_t$ added at each timestep $t$, conditioned on text $c$.
Following DDPM~\cite{ho2020denoising}, the training objective simplifies the variational lower bound of $\mathbb E_q[-\log p_\theta(x_0)]$ into prior matching term $L_T$, denoising matching terms $L_{t-1}$, and reconstruction term $L_0$:
\begin{align}
    &L = \mathbb{E}_q\big[
    \underbrace{D_{KL}\big(q(x_T|x_0)||p(x_T)\big)}_{L_T} + \\
    &\sum_{t>1} \underbrace{D_{KL}\big(q(x_{t-1}|x_t, x_0)||p_{\theta}(x_{t-1}|x_t)\big)}_{L_{t-1}}
    \underbrace{- \log p_\theta(x_0|x_1)}_{L_0}
    \big].\nonumber 
\end{align}
where $p_\theta$ denotes the learned data distribution, implicitly defined by the reverse process parameterized by the noise prediction network $\epsilon_\theta$.
In practice, this objective is normally simplified to a reweighted MSE loss~\cite{ho2020denoising}, which serves as a simpler and widely adopted target:
\begin{align}
    L:=\mathbb{E}_{t,x_0,\epsilon}\big[
    w(t)||\epsilon_t - \epsilon_\theta(x_t,t, c)||^2
    \big],
\end{align}
where $t$ is uniformly sampled from $\{1, \dots, T\}$, $\epsilon_t \sim \mathcal{N}(0, \mathbf{I})$, and $x_t$ is constructed via the forward process.
The weight $w(t)$ is typically set to a constant.
Critically, training depends on the predefined forward noise schedule.
Despite this dependency, once trained, the same model $\epsilon_\theta$ can be used with various sampling schedulers (e.g., DDPM~\cite{ho2020denoising}, DDIM~\cite{song2020denoising}) at inference time without retraining.

At inference time, image generation begins from pure Gaussian noise $x_T \sim \mathcal{N}(0, \mathbf{I})$ and iteratively denoises it toward a clean latent $\hat{x}_0$.
A general reverse update rule can be written as~\cite{sun2025noise}:
\begin{align}
x_{t-1} = \kappa_t x_t + \lambda_t \epsilon_\theta(x_t, t, c) + \eta_t \tilde{\epsilon}_t,
\label{eq:reverse_origin}
\end{align}
where $\epsilon_\theta(x_t, t, c)$ is the predicted noise at step $t$, $\tilde{\epsilon}_t \sim \mathcal{N}(0, \mathbf{I})$ is newly sampled noise, $\kappa_t$, $\lambda_t$, and $\eta_t$ are coefficients derived from both the forward noise schedule and the chosen sampling algorithm.
After $T$ reverse steps, the latent $\hat{x}_0$ is decoded into an image $\hat v=\mathcal{D}(\hat{x}_0)$. Through this process, LDMs generate high-quality images.

\section{Method}

\begin{figure}[!t]
    \centering
    \includegraphics[width=0.9\linewidth]{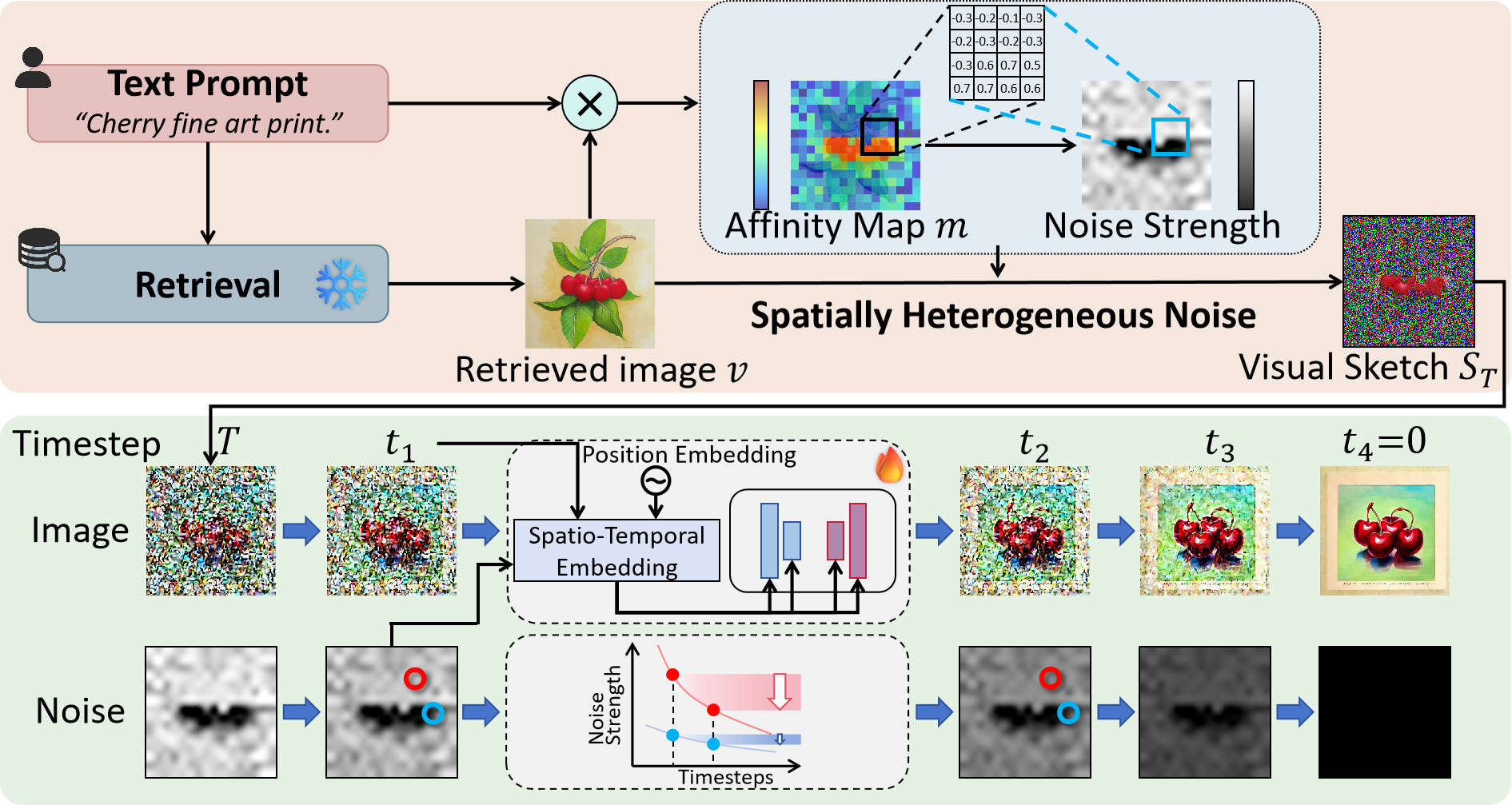}
    \caption{Overview of PixARG. Based on the semantic relevance between the retrieved image $v$ and the text prompt, PixARG selectively applies spatially heterogeneous noise to obtain a visual sketch $S_T$ as the initial latent for generation. Text-aligned regions are perturbed with weaker noise to preserve structural details, while less relevant areas receive stronger noise to encourage regeneration. During sampling, pixel-wise asynchronous denoising allows pixels with different initial noise levels to denoise at adaptive rates, completing the generation within the same steps.}
    \label{overview}
\end{figure}

We propose a pixel-wise asynchronous retrieval-augmented generation (PixARG) framework that integrates retrieval into the diffusion process to enable both fast and photorealistic image synthesis, as shown in Fig. \ref{overview}.
Given a text prompt, PixARG first retrieves a reference image from a pre-constructed retrieval database, and simultaneously derives a pixel-wise text-visual affinity map, quantifying how well each spatial location matches the prompt.
To incorporate this retrieved content into the diffusion pipeline as visual prior, we construct a visual sketch by selectively adding noise to the reference image, where the strength of noise applied to each pixel is decided by the affinity map. Then, the visual sketch serves as the initial latent for the reverse diffusion process, where different regions denoise at different rates according to their affinity map and noise levels.
To support this spatially asynchronous diffusion mechanism, we modify the denoising network by conditioning it jointly on the timestep and the affinity map, allowing the model to adapt its denoising behavior spatially. Furthermore, we reformulate the training objective to accommodate this non-uniform generative process, encouraging the model to balance deterministic restoration and stochastic exploration according to spatial alignment.

\subsection{Retrieving Image for Text-Visual Affinity Mapping}

\begin{comment}
\begin{align}
v = \text{Retrieve}(\mathcal{V} \mid c),
\end{align}
\end{comment}

A candidate image library $\mathcal{V}$ for retrieval is first constructed from user-provided or web-sourced images, curated to ensure broad semantic and visual diversity. Given a text prompt $c$, a cross-modal retriever, such as a vision-language model or commercial image search engine, selects an image $v \in \mathcal{V}$ as the visual prior, which is semantically aligned with the input text prompt.

As different regions of the retrieved image align with the text prompt to varying degrees, it is necessary to distinguish meaningful guidance from noise. We thus compute a pixel-level text-visual affinity map $m \in [-1,1]^{h\times w}$ over the retrieved image $v$. Each element $m^{(i)}$ quantifies the semantic alignment between the text prompt $c$ and the image patch at spatial position $i$, which allows for suppressing irrelevant regions during spatially-aware conditioning. Formally, let $f_I$ and $f_T$ denote the image and text encoders, respectively. Then, the affinity map is derived from the cosine similarity between the visual features of each patch and the text embedding: 
\begin{align}
m^{(i)} = \frac{ f_I(v)^{(i)\top} f_T(c) }{ \lVert f_I(v)^{(i)} \rVert \cdot \lVert f_T(c) \rVert },
\end{align}
for $i\in\{1,2,\dots,h\times w\}$. By filtering and reweighing the retrieved visual prior, the affinity map ensures that only semantically aligned regions contribute to the diffusion process. This allows the model to leverage the prior's fine details while simultaneously avoiding semantic noise from unrelated regions.

\subsection{Affinity-Guided Latent Noise Modulation}

To leverage the retrieved image to provide composition, texture, and structural cues, we encode it as a visual prior in the latent space. Given that its semantic relevance to text prompt is spatially non-uniform, we use the affinity map $m$ to selectively guide the forward process. Rather than applying uniform noise across the image, we reformulate the diffusion process by spatially modulating the noise intensity according to $m$. Regions with high semantic relevance receive weaker noise perturbations, thereby preserving fine visual details. Conversely, less relevant areas are injected with stronger noise, effectively erasing inconsistent details and enabling the free synthesis of content, guided primarily by the text prompt.

Building upon this principle, we define the latent representation at timestep $t \in \{1,\dots,T\}$ as a visual sketch $S_t$, which encodes pixel-wise, heterogeneous noise intensities modulated by $m$:
\begin{align}
    S_t &= a_{t,m} \odot x_0 + b_{t,m} \odot \epsilon_t,
    \label{eq:sketch}
\end{align}
where $a_{t,m}, b_{t,m} \in [0,1]^{h\times w}$ are spatially adaptive coefficients derived from the base noise schedules $\{a_t\}_{t=1}^T$ and $\{b_t\}_{t=1}^T$, reweighted by $t$ and $m$; and $\odot$ denotes element-wise multiplication. Such a mechanism associates the reliable areas of the visual sketch with the retrieved prior, while allowing the irrelevant regions to reset for regeneration. As a result, semantic consistency is ensured without sacrificing model flexibility. Compared with feature-level injection techniques, our latent-space design preserves richer spatial and textural information, yielding better visual fidelity and efficiency.

\subsection{Pixel-Wise Asynchronous Denoising}

We leverage the retrieved visual information by initializing the reverse denoising process from the visual sketch $S_T$, which, unlike standard Gaussian noise, contains spatially varying noise levels. To reconcile this spatial heterogeneity with the globally shared timestep of the standard diffusion framework, we propose a pixel-wise asynchronous denoising scheduler, which dynamically adjusts the denoising trajectory for each spatial location based on its initial noise intensity, such that all pixels converge to a spatially coherent output after the same number of denoising steps.

Specifically, we define spatially-dependent coefficients $\kappa_{t,m}$, $\lambda_{t,m}$, and $\eta_{t,m}$, which rescale the denoising dynamics according to the affinity map $m$ and timestep $t$. These coefficients modulate the contributions of the current latent, predicted noise, and stochastic term, respectively, yielding spatially adaptive reverse process:
\begin{align}
\begin{aligned}
    x_{t-1} := &\kappa_{t,m} \odot x_t + \lambda_{t,m} \odot \epsilon_\theta(x_t, t, m, c) + \eta_{t,m} \odot \tilde\epsilon_t,
\end{aligned}
    \label{eq:noise_pred}
\end{align}
where $\epsilon_\theta(x_t, t, m, c)$ denotes the noise predicted by the diffusion model $\theta$ conditioned on the prompt $c$ and guided by the spatial text-visual affinity map $m$, and $\tilde\epsilon_t$ is sampled from a standard Gaussian distribution. To ensure consistency with the forward process that generated $S_T$, the coefficients $\kappa_{t,m}$ and $\lambda_{t,m}$ are constrained by the spatially adaptive cumulative noise schedules $a_{t,m}$ and $b_{t,m}$:
\begin{align}
\begin{aligned}
    \kappa_{t,m}&=\frac{a_{t-1,m}}{a_{t,m}},\\
    \lambda_{t,m}&=-\frac{a_{t-1,m}\odot b_{t,m}}{a_{t,m}} + \sqrt{b^2_{t-1,m}-\eta^2_{t,m}}.
\end{aligned}\label{eq:global_constraint}
\end{align}
Initializing from $S_T$ also confers significant computational and perceptual advantages. In regions where the retrieved image is semantically aligned with the text prompt, the latent state begins much closer to the data manifold than random Gaussian noise. Although all spatial locations undergo the same number of denoising steps, these well-initialized regions require fewer corrective updates during generation, leading to more stable convergence and earlier emergence of high-fidelity structures. As a result, each denoising step is more effective, enhancing output realism and structural coherence without requiring additional computational steps.

\subsection{Spatiotemporal Denoising Network}

To accommodate the spatially asynchronous denoising process, we propose a spatiotemporal denoiser $\epsilon_\theta(x_t, t, m, c)$, extending the standard diffusion network by explicitly incorporating the text-visual affinity map $m$ into noise prediction. Unlike conventional diffusion models where the timestep $t$ alone controls a uniform noise schedule, our method determines the noise intensity at the pixel level jointly from $t$ and $m$, where $t$ controls the global denoising trajectory, while $m$ serves as a spatial modulator, allowing each pixel's denoising in an asynchronous way. 

From implementation, $m$ is passed through a lightweight convolutional or linear projection module, then summed with both timestep and position embeddings at the pixel level to create a spatiotemporal embedding. This joint conditioning scheme preserves spatial correspondence, allowing the network to selectively focus on region-specific details while maintaining global coherence. Consequently, the network can benefit from location-aware guidance for selective restoration and resynthesis, while remaining fully compatible with standard diffusion model architectures.

\subsection{Loss Function and Training Process}\label{sec:loss}

Our pixel-wise asynchronous scheduler applies spatially varying noise schedules to individual pixels, unlike conventional DDPMs that employ a globally shared, spatially uniform schedule. This shift renders standard loss incompatible, necessitating a re-derivation of the core timestep loss $L_t$ under a per-pixel parameterization.

Assuming the true posterior $q(x_{t-1}|x_t,x_0)$ and the model distribution $p_\theta(x_{t-1}|x_t)$ are diagonal Gaussian when $\eta\neq0$ (i.e., $q\sim\mathcal{N}(\mu_q,\Sigma_q)$ and $p_\theta\sim\mathcal{N}(\mu_p,\Sigma_p)$), with diagonal covariances $\Sigma_q$ and $\Sigma_p$, the KL divergence reduces to a sum of pixel-wise terms:
\begin{align}
\begin{aligned}
    &KL\left(q||p_\theta\right) = \frac{1}{2}\Big[\text{tr}(\Sigma_p^{-1}\Sigma_q)\\
    &+(\mu_p-\mu_q)^T\Sigma_p^{-1}(\mu_p-\mu_q)
    -h\times w+\ln\frac{\text{det}\Sigma_p}{\text{det}\Sigma_q}\Big]\\
    &= \sum_{i}\frac{1}{2}\Big[\frac{\Sigma_q^{(i)}}{\Sigma_p^{(i)}}+\frac{(\mu_p^{(i)}-\mu_{q}^{(i)})^2}{\Sigma_p^{(i)}}-1+\ln\frac{\Sigma_p^{(i)}}{\Sigma_q^{(i)}}\Big].
\end{aligned}
\end{align}

\begin{table*}[t]
\centering
\caption{Quantitative results on the SD 1.4 backbone under DDIM and PNDM samplers on the LAION dataset. Inference time measures the total duration from prompt to image, encompassing all stages such as retrieval, encoding, and generation.}
\renewcommand{\arraystretch}{0.8}
\resizebox{\linewidth}{!}{
\begin{tabular}{lccccccccccccccc}
\toprule
\multirow{2}{*}{Method} & 
\multirow{2}{*}{\shortstack{Diffusion\\Steps↓}} & 
\multirow{2}{*}{\shortstack{Time\\(s/img)↓}} &
\multicolumn{2}{c}{Basic Metrics} & 
\multicolumn{2}{c}{Realism \& Diversity} & 
\multicolumn{4}{c}{RichHF-18K} & 
\multicolumn{3}{c}{Human Preference Metrics} \\
\cmidrule(lr){4-5} \cmidrule(lr){6-7} \cmidrule(lr){8-11} \cmidrule(lr){12-14}
& & & CLIP↑ & FID↓ & Precision(\%)↑ & Recall(\%)↑ & Plausibility↑ & Alignment↑ & Aesthetics↑ & Overall↑ & HPSv2↑ & MPS↑ & PickScore↑ \\ 
\midrule
\multicolumn{14}{l}{\textit{With DDIM~\cite{song2020denoising} Sampler}} \\
SD 1.4~\cite{sdv1} & 50 & 0.883 & 33.10 & 11.49 & 57.07 & \textbf{64.73} & 0.6892 & 0.6079 & \textbf{0.6753} & 0.5828 & 24.00 & 9.58 & 19.85 \\
ReDi~\cite{zhang2023redi} & 40 & 0.707 & 26.71 & 13.87 & 58.71 & 44.46 & 0.6589 & 0.5978 & 0.6488 & 0.5646 & 21.22 & 8.23 & 18.84 \\
DiT4Edit~\cite{feng2025dit4edit} & 50 & 2.767 & 30.92 & 9.44 & 64.34 & 59.65 & 0.6800 & 0.6008 & 0.6586 & 0.5739 & 21.86 & 9.18 & 19.47 \\
RDM~\cite{blattmann2022retrieval} & 50 & 0.890 & 32.98 & 11.31 & 58.40 & 61.30 & 0.6871 & 0.6075 & 0.6724 & 0.5815 & 24.01 & 9.54 & 19.82 \\
KNN-Diffusion~\cite{sheynin2022knn} & 50 & 0.869 & 32.55 & 13.14 & 55.30 & 60.76 & 0.6859 & 0.6081 & 0.6690 & 0.5812 & 23.46 & 9.20 & 19.53 \\
Re-Imagen~\cite{chen2022re} & 50 & 1.313 & 32.71 & 11.09 & 58.24 & 61.98 & 0.6892 & 0.6052 & 0.6678 & 0.5795 & 23.37 & 9.24 & 19.65 \\
\rowcolor{gray!10}PixARG (Ours) & \textbf{15} & \textbf{0.367} & \textbf{33.43} & \textbf{6.78} & \textbf{69.01} & 61.35 & \textbf{0.6964} & \textbf{0.6103} & 0.6726 & \textbf{0.5860} & \textbf{24.40} & \textbf{9.96} & \textbf{20.06} \\
\midrule
\multicolumn{14}{l}{\textit{With PNDM~\cite{liu2022pseudo} Sampler}} \\
SD 1.4~\cite{sdv1} & 50 & 0.883 & 33.06 & 11.66 & 55.65 & \textbf{65.66} & 0.6924 & 0.6083 & 0.6763 & 0.5842 & 24.05 & 9.51 & 19.81 \\
ReDi~\cite{zhang2023redi} & 40 & 0.707 & 29.66 & 15.94 & 55.09 & 41.84 & 0.6568 & 0.5950 & 0.6455 & 0.5626 & 20.85 & 7.98 & 18.67 \\
DiT4Edit~\cite{feng2025dit4edit} & 50 & 2.767 & 33.22 & 9.70 & 63.60 & 61.74 & 0.6830 & 0.6064 & 0.6702 & 0.5804 & 24.40 & 9.56 & 19.99 \\
RDM~\cite{blattmann2022retrieval} & 50 & 0.890 & 33.09 & 10.17 & 56.67 & 61.34 & 0.6988 & 0.6119 & 0.6812 & 0.5893 & 24.35 & 9.57 & 19.87 \\
KNN-Diffusion~\cite{sheynin2022knn} & 50 & 0.869 & 32.53 & 13.45 & 53.57 & 60.64 & 0.6889 & 0.6084 & 0.6700 & 0.5825 & 23.56 & 9.20 & 19.51 \\
Re-Imagen~\cite{chen2022re} & 50 & 1.313 & 32.70 & 11.21 & 56.88 & 61.59 & 0.6973 & 0.6079 & 0.6770 & 0.5855 & 23.52 & 9.18 & 19.62 \\
\rowcolor{gray!10}PixARG (Ours) & \textbf{15} & \textbf{0.367} & \textbf{33.28} & \textbf{7.12} & \textbf{65.69} & 61.44 & \textbf{0.7084} & \textbf{0.6154} & \textbf{0.6864} & \textbf{0.5945} & \textbf{24.72} & \textbf{9.62} & \textbf{20.03} \\
\bottomrule
\end{tabular}}
\label{tab:main_results}
\end{table*}

Under our asynchronous scheduler, both $\Sigma_q$ and $\Sigma_p$ become pixel-wise functions of the affinity map $m$. The entropy and cross-entropy terms are absorbed into a time- and affinity-dependent constant $C(t,m)$, leaving the mean-matching term as the primary focus of optimization. The effective loss thus simplifies to:
\begin{align}
    L = \mathbb{E}_{c,\epsilon, t} \mathbb{E}_{m,v} \sum_i \Big( \frac{(\mu_p^{(i)}-\mu_{q}^{(i)})^2}{\Sigma_p^{(i)}} + C(t,m) \Big),
% &= \mathbb{E}_{c,\epsilon, t,m,v} \sum^{(i)} \big( \frac{(\mu_{\theta}^{(i)}-\mu_{q}^{(i)})^2}{\Sigma_\theta^{(i)}} + C(t,m) \big) \\
    \label{eq:loss_mu+sigma}
\end{align}
which can be expressed via noise prediction error:
\begin{align}
    L \propto \mathbb{E}_{c,\epsilon, t,m,v} \sum_i \Big(\frac{\lambda_{t,m}^{(i)}}{\eta_{t,m}^{(i)}}\Big)^2(\epsilon_{\theta}^{(i)}-\epsilon_t^{(i)})^2,
    \label{eq:loss_epsilon}
\end{align}
where $\lambda$ and $\eta$ are the scaling coefficients for the predicted noise $\epsilon_\theta$ and the random noise in Eq. \ref{eq:noise_pred}, respectively, and $\epsilon_t^{(i)}$ is the added noise at pixel $i$.

During training, the affinity map $m$ is sampled from a component Gaussian Mixture Model that approximates the distribution of real text-visual affinity maps, providing diverse training signals to enhance robustness. This enables training under spatially adaptive diffusion dynamics while preserving the denoising objective.

\section{Experiment}

\subsection{Experimental Setup}

\begin{table*}[t]
\centering
\caption{Quantitative comparison on the SD 1.4 backbone under DDIM sampler on the COYO dataset.}
\renewcommand{\arraystretch}{0.75}
\resizebox{\linewidth}{!}{
\begin{tabular}{lcccccccccccc}
\toprule
\multirow{2}{*}{Method} & \multirow{2}{*}{Steps↓} & 
\multicolumn{2}{c}{Basic Metrics} & 
\multicolumn{2}{c}{Realism \& Diversity} & 
\multicolumn{4}{c}{RichHF-18K} & 
\multicolumn{3}{c}{Human Preference Metrics} \\
\cmidrule(lr){3-4} \cmidrule(lr){5-6} \cmidrule(lr){7-10} \cmidrule(lr){11-13}
& & CLIP↑ & FID↓ & Precision(\%)↑ & Recall(\%)↑ & Plausibility↑ & Alignment↑ & Aesthetics↑ & Overall↑ & HPSv2↑ & MPS↑ & PickScore↑ \\ 
\midrule
SD 1.4~\cite{sdv1} & 50 & 33.77 & 9.33 & 64.35 & 70.44 & 0.5994 & 0.5502 & 0.6028 & 0.5189 & \textbf{22.90} & 8.57 & 19.58 \\
ReDi~\cite{zhang2023redi} & 40 & 27.63 & 14.57 & 61.86 & 49.29 & 0.5398 & 0.5337 & 0.5582 & 0.4860 & 18.08 & 6.34 & 17.96 \\
DiT4Edit~\cite{feng2025dit4edit} & 50 & 28.15 & 9.20 & 68.11 & 64.89 & 0.5953 & 0.5504 & 0.5998 & 0.5168 & 19.11 & 7.14 & 18.45 \\
RDM~\cite{blattmann2022retrieval} & 50 & 33.23 & 11.90 & 63.21 & 69.83 & 0.6073 & 0.5501 & \textbf{0.6059} & 0.5218 & 22.82 & 8.40 & 19.47 \\
KNN-Diffusion~\cite{sheynin2022knn} & 50 & 33.16 & 10.99 & 62.52 & 69.03 & 0.6039 & 0.5480 & 0.6043 & 0.5195 & 21.40 & 8.29 & 19.35\\
Re-Imagen~\cite{chen2022re} & 50 & 32.53 & 12.81 & 64.50 & 65.93 & 0.6029 & 0.5441 & 0.5493 & 0.5148 & 21.22 & 7.74 & 19.06 \\
\rowcolor{gray!10}PixARG (Ours) & \textbf{15} & \textbf{33.79} & \textbf{8.23} & \textbf{73.47} & \textbf{70.65} & \textbf{0.6205} & \textbf{0.5518} & 0.6058 & \textbf{0.5272} & 22.43 & \textbf{8.69} & \textbf{19.65} \\
\bottomrule
\end{tabular}}
\label{tab:main_results_coyo}
\end{table*}

\textbf{Dataset.} For offline experiments, we follow~\cite{luo2023latent,ye2025schedule,huang2025illume+} and use LAION-2B-en~\cite{schuhmann2022laion} and COYO-700M~\cite{kakaobrain2022coyo-700m} for 512×512 resolution experiments, and relaion-high-resolution~\cite{schuhmann2022laion} for 1024×1024 resolution.
We randomly sample 80k image–text pairs for training, 20k for evaluation, and another 20k for the retrieval database.

For offline user studies, we use RedCaps~\cite{desai2021redcaps} with author-annotated image–text pairs. To simulate personalized retrieval, we perform per-user evaluation: 20\% of each author's submissions act as prompts, while the rest form their private retrieval database.

For online experiments, we use LAION as the prompt source and Google Image Search\footnote{\url{https://images.google.com/}} as the online retriever.

\noindent
\textbf{Text-to-Image Models.} We integrate our method into several representative diffusion models.
Our main baseline is Stable Diffusion 1.4 (SD 1.4)~\cite{sdv1}, a UNet-based model generating 512×512 images.
We also evaluate PixArt-$\alpha$~\cite{chen2023pixart}, a diffusion Transformer using its 512×512 version, and SDXL~\cite{podell2023sdxl}, a UNet-based model generating 1024×1024 images.
Each model uses its default sampler for fairness: DDIM~\cite{song2020denoising} and PNDM~\cite{liu2022pseudo} for SD 1.4, DPM-Solver++~\cite{lu2025dpm} for PixArt-$\alpha$, and Euler Discrete~\cite{karras2022elucidating} for SDXL.

\noindent
\textbf{Retrieval Method.} We employ two complementary retrieval strategies: TextSpan~\cite{gandelsman2023interpreting} for offline retrieval and Google Image Search for online retrieval. TextSpan computes pixel-level similarity between prompts and images to derive text-visual affinity maps.

\noindent
\textbf{Baselines.} We use the original Stable Diffusion model as our primary baseline.
For comparison, we also adapt ReDi~\cite{zhang2023redi}, DiT4Edit~\cite{feng2025dit4edit}, RDM~\cite{blattmann2022retrieval}, KNN-Diffusion~\cite{sheynin2022knn}, and Re-Imagen~\cite{chen2022re} to our setting.
ReDi accelerates generation by reusing previously generated diffusion trajectories.
DiT4Edit, originally developed for image editing, is adapted as an image-to-image baseline combined with an image retriever.
RDM, KNN-Diffusion and Re-Imagen retrieves visual examples to synthesize unseen concepts. We adapt them to the stable diffusion framework for fair comparison.

\noindent
\textbf{Metrics.}
We evaluate alignment, realism, and visual quality using multiple metrics.
CLIP~\cite{radford2021learning} measures text–image alignment; FID~\cite{heusel2017gans} evaluates distributional similarity to real images.
Precision and Recall~\cite{kynkaanniemi2019improved} capture realism and diversity.
RichHF-18K benchmark~\cite{liang2024rich} is adopted to assess plausibility, alignment, and aesthetics.
Human preference metrics, such as HPSv2~\cite{wu2023human}, MPS~\cite{zhang2024learning}, and PickScore~\cite{kirstain2023pick}, are also included for aesthetic evaluation.

\noindent
\textbf{Training Configurations.}
We fine-tune diffusion model using a rank-4 LoRA~\cite{hu2022lora} for 100k steps with batch size 16. Learning rates are set to $5\times10^{-5}$ for the UNet and $1\times10^{-5}$ for the Transformer.

\subsection{Main Results}

\begin{figure}[!t]
    \centering
    \includegraphics[width=0.85\linewidth]{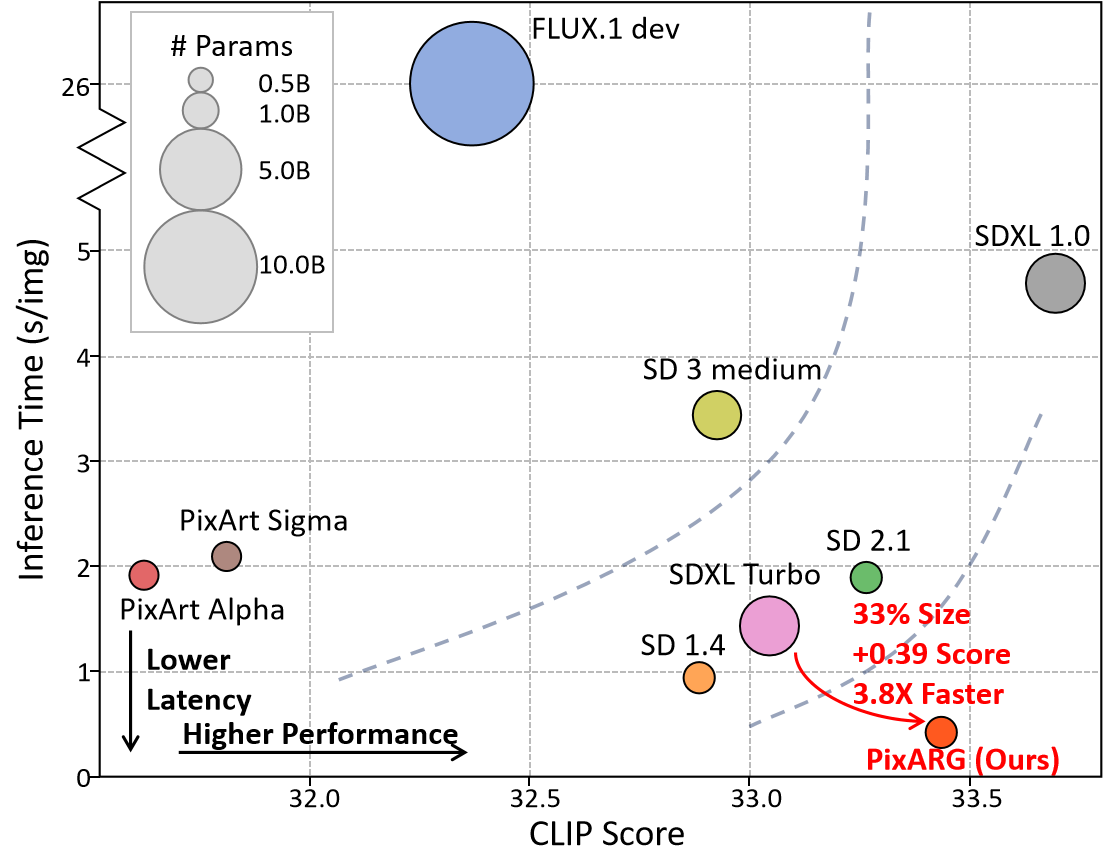}
    \caption{Performance comparison of diffusion models with varying parameter scales. Combined with PixARG, the smaller SD 1.4 outperforms SDXL Turbo.}
    \label{fig:compare-to-larger-model}
\end{figure}

We present quantitative results on the LAION dataset in Tab. \ref{tab:main_results} and the results on the COYO dataset in Tab. \ref{tab:main_results_coyo}.
By leveraging retrieved real images as visual priors, PixARG achieves superior performance with only 15 diffusion steps, surpassing the SD~1.4 baseline that requires 50 steps.
Notably, PixARG yields the largest improvement in CLIP score by 0.33 and 0.22 on LAION dataset, demonstrating that semantically aligned priors effectively enhance text–image correspondence.
Precision and Plausibility also increase by 11.94\% and 1.04\% on LAION, and by 9.12\% and 3.52\% on COYO, indicating that incorporating real visual priors leads to more realistic and physically consistent generations.
Recall slightly decreases on LAION but not on COYO, attributed to differences in retrieval database quality, e.g., whether similar prompts retrieve redundant versus diverse images.
Results on human-preference metrics further show that PixARG’s generations exhibit stronger aesthetic agreement with human judgment.
In contrast, other methods perform less effectively under the same setting.
RDM, which exploits only semantic information from retrieved images, produces results comparable to SD~1.4.
ReDi, which accelerates sampling by reusing diffusion trajectories, underperforms slightly since trajectory reuse does not expand the model’s generative capacity.
Overall, PixARG uniquely integrates retrieved images as visual priors to improve both generation efficiency and perceptual quality simultaneously.

We further compare PixARG based on SD 1.4 with diffusion models of varying sizes in Fig. \ref{fig:compare-to-larger-model}.
With the same parameter size, PixARG improves the CLIP score by 0.54 and achieves 2.4× inference speedup over SD 1.4. Additionally, with only 1/3 of the parameters of SDXL 1.0, PixARG achieves comparable CLIP performance and is 12.6× faster.
This shows that our retrieval-augmented diffusion framework enables smaller models to reach the performance of much larger ones while operating substantially faster.

\begin{figure}[t!]
    \centering
    \begin{subfigure}[b]{\linewidth}
        \centering
        \includegraphics[width=0.9\linewidth]{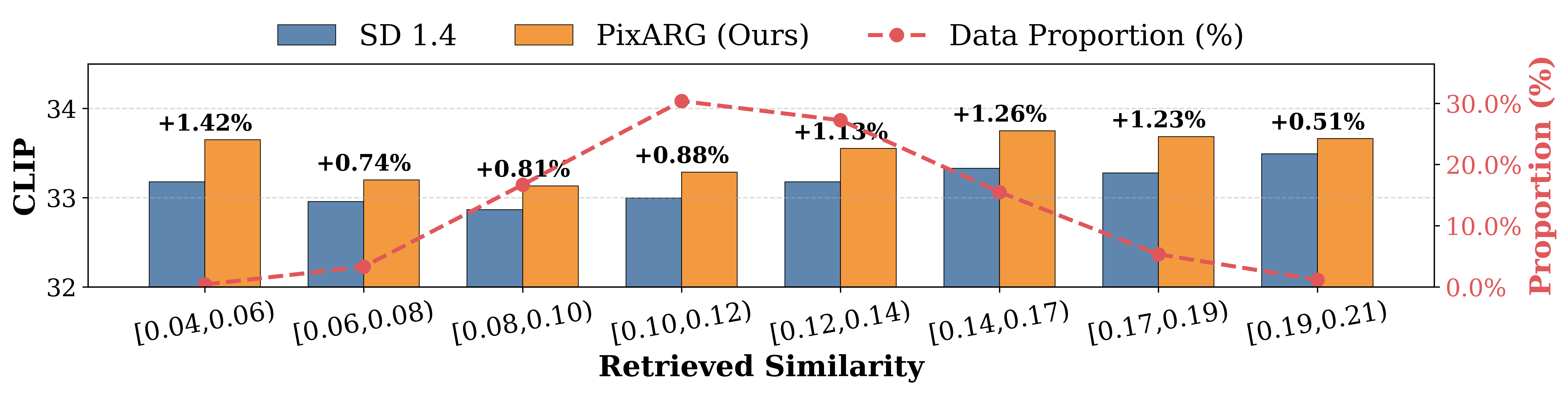}
        \caption{Semantic alignment (CLIP score).}
    \end{subfigure}
    \hfill
    \begin{subfigure}[b]{\linewidth}
        \centering
        \includegraphics[width=0.9\linewidth]{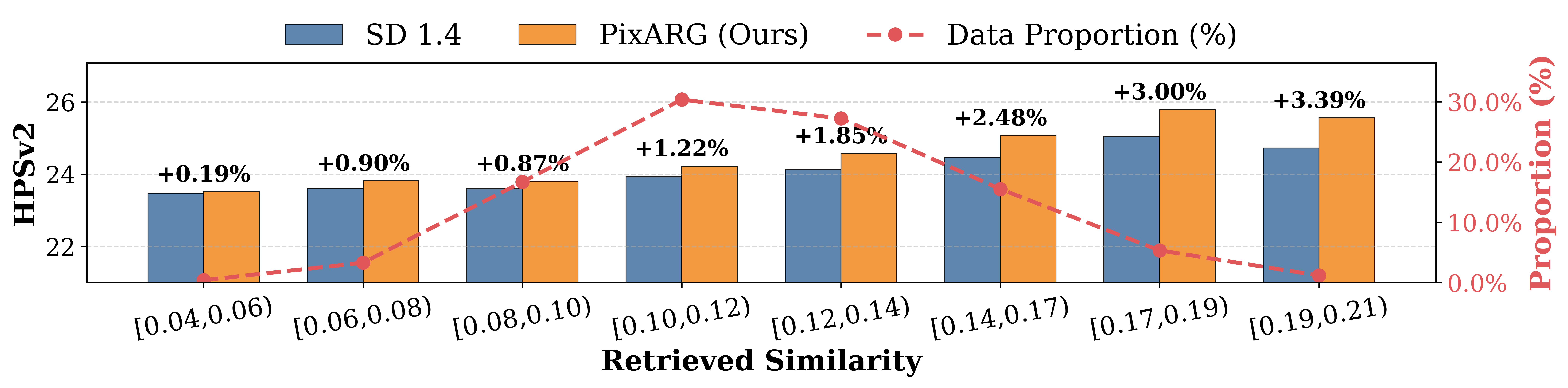} 
        \caption{Human preference (HPSv2).}
    \end{subfigure}
    \hfill
    \begin{subfigure}[b]{\linewidth}
        \centering
        \includegraphics[width=0.9\linewidth]{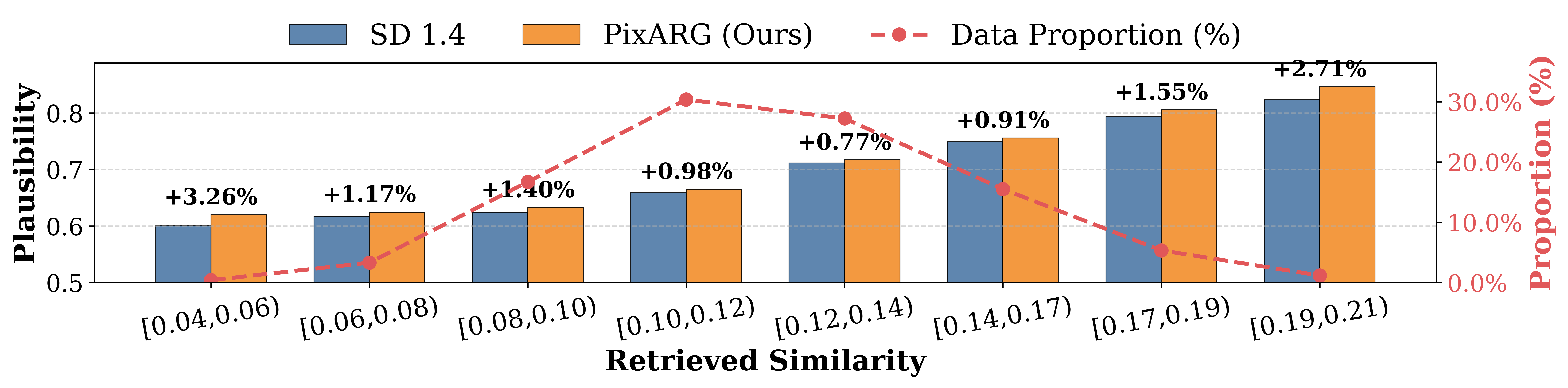}
        \caption{Visual realism (RichHF-18K Plausibility).}
    \end{subfigure}

    \caption{Our PixARG achieves consistent performance gains across all retrieval similarity intervals, with metrics evaluating alignment, human preference, and realism.
    }
    \label{fig:re1}
\end{figure}

\subsection{Sensitivity Analysis on Retrieval Quality}

To evaluate the sensitivity of our PixARG to retrieval quality, we analyze its performance across various retrieval similarity intervals in Fig. \ref{fig:re1}. Our method demonstrates consistent gains over the SD1.4 baseline across all intervals. Notably, even in extreme low-similarity regions ($<$0.06), which typically indicate retrieval failures such as missing matches or inaccurate affinity maps, our method still improves alignment by 1.42\% and realism by 3.26\%. This robustness is primarily attributed to our similarity-aware noise mechanism, which effectively suppresses unreliable, perturbation-sensitive semantic cues while preserving essential low-level image statistics. 
Consequently, by preserving structural and frequency priors inherent in natural images, this structured initialization outperforms standard isotropic Gaussian noise, ensuring stable generation performance even under high retrieval noise.

\subsection{Ablation Study}

\textbf{PixARG Components Ablation.} Tab. \ref{tab:component_ablation} details the component ablation study. The \textit{Retrieval} baseline bypasses generation to directly output retrieved images; while this ensures high realism and diversity, it suffers from poor text–image alignment due to prompt mismatches. Using synchronous denoising also degrades CLIP alignment, as the model fails to distinguish between semantically matched and unmatched regions. Conversely, employing asynchronous denoising with a standard MSE loss leverages heterogeneous noise to filter irrelevant regions, yielding a CLIP improvement over synchronous denoising. This suggests that heterogeneous noise helps focus on relevant visual priors. However, without our spatially adaptive loss, the model's capacity to handle spatial heterogeneity remains constrained, limiting precision. Ultimately, the full PixARG model achieves the optimal balance between realism and alignment, validating the effectiveness of both retrieval guidance and asynchronous denoising.

\begin{table}[!t]
\centering
\caption{Ablation study under the DDIM sampler on the LAION dataset, using 15 diffusion steps for all settings. ``Ret.'' denotes the Rank-1 retrieved image without diffusion; ``Sync.'' adds synchronous denoising; ``MSE'' replaces the reconstruction objective with MSE loss; and ``Rank-$k$'' variants take the $k$-th retrieved image instead of the default Rank-1 image.}
\resizebox{\linewidth}{!}{
\begin{tabular}{lccccccccc}
\toprule
 &
\multicolumn{2}{c}{Basic Metrics} &
\multicolumn{2}{c}{Realism \& Diversity} &
\multicolumn{4}{c}{RichHF-18K} \\
\cmidrule(lr){2-3} \cmidrule(lr){4-5} \cmidrule(lr){6-9}
& CLIP↑ & FID↓ & Precision(\%)↑ & Recall(\%)↑ & Plausibility↑ & Alignment↑ & Aesthetics↑ & Overall↑ \\ 
\midrule
Ret. & 32.34 & 4.40 & 76.68 & 69.24 & 0.7145 & 0.6017 & 0.6700 & 0.5887 \\
\midrule
Sync. & 33.30 & 8.64 & \textbf{69.31} & 59.49 & 0.6869 & 0.6030 & 0.6637 & 0.5783 \\
MSE & \underline{33.38} & \underline{6.94} & 66.84 & \underline{61.30} & \underline{0.6958} & \underline{0.6093} & \textbf{0.6734} & \underline{0.5852} \\
\rowcolor{gray!10}Ours & \textbf{33.43} & \textbf{6.78} & \underline{69.01} & \textbf{61.35} & \textbf{0.6964} & \textbf{0.6103} & \underline{0.6726} & \textbf{0.5860} \\
\midrule
Rank-2 & 33.35 & 6.98 & 69.13 & 59.78 & 0.6973 & 0.6098 & 0.6722 & 0.5859 \\
Rank-3 & 33.32 & 7.00 & 69.88 & 60.72 & 0.6980 & 0.6107 & 0.6733 & 0.5868 \\
Rank-5 & 33.27 & 6.99 & 69.57 & 59.56 & 0.6966 & 0.6095 & 0.6712 & 0.5854 \\
\bottomrule
\end{tabular}}
\label{tab:component_ablation}
\end{table}

\begin{figure}[t!]
    \centering
    \begin{subfigure}[b]{0.3\linewidth}
        \centering
        \includegraphics[width=\linewidth]{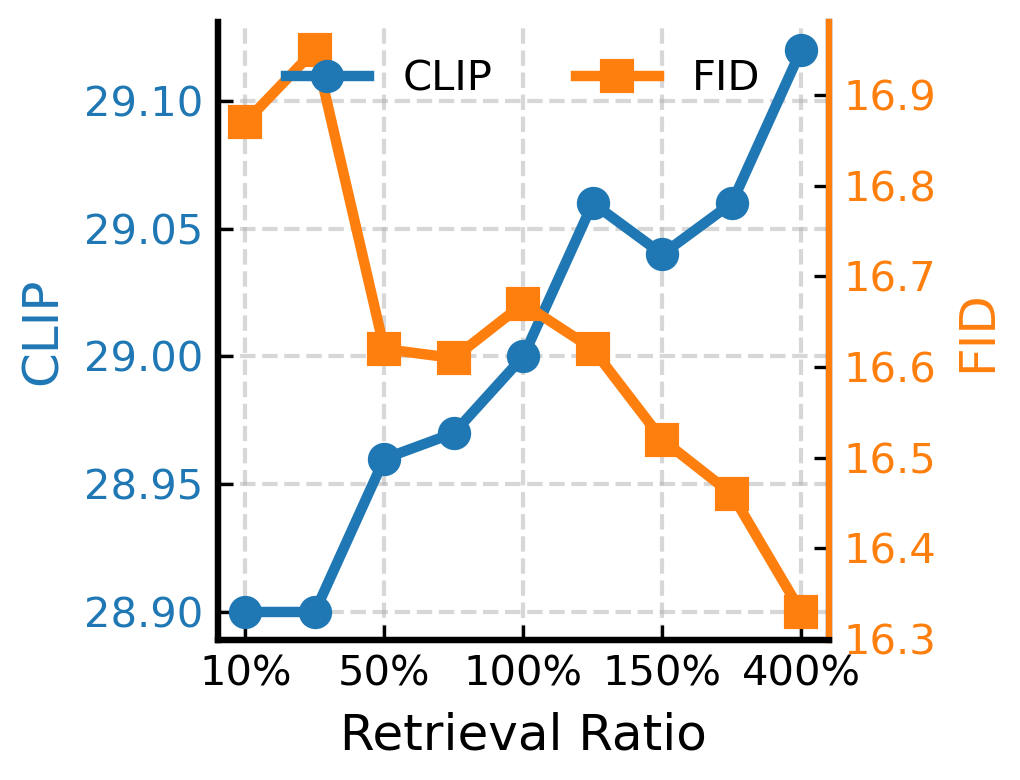}
        \caption{CLIP \& FID}
        \label{fig:ratio-clip-fid}
    \end{subfigure}
    \hfill
    \begin{subfigure}[b]{0.32\linewidth}
        \centering
        \includegraphics[width=\linewidth]{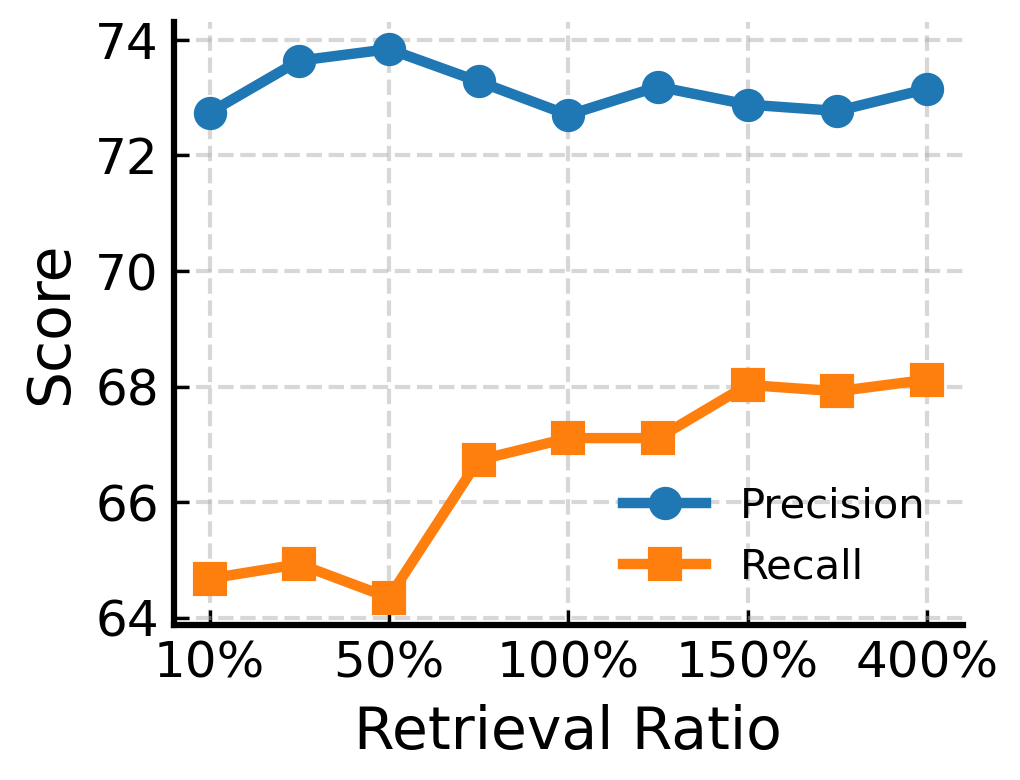}
        \caption{Precision \& Recall}
        \label{fig:ratio-pr}
    \end{subfigure}
    \hfill
    \begin{subfigure}[b]{0.3\linewidth}
        \centering
        \includegraphics[width=\linewidth]{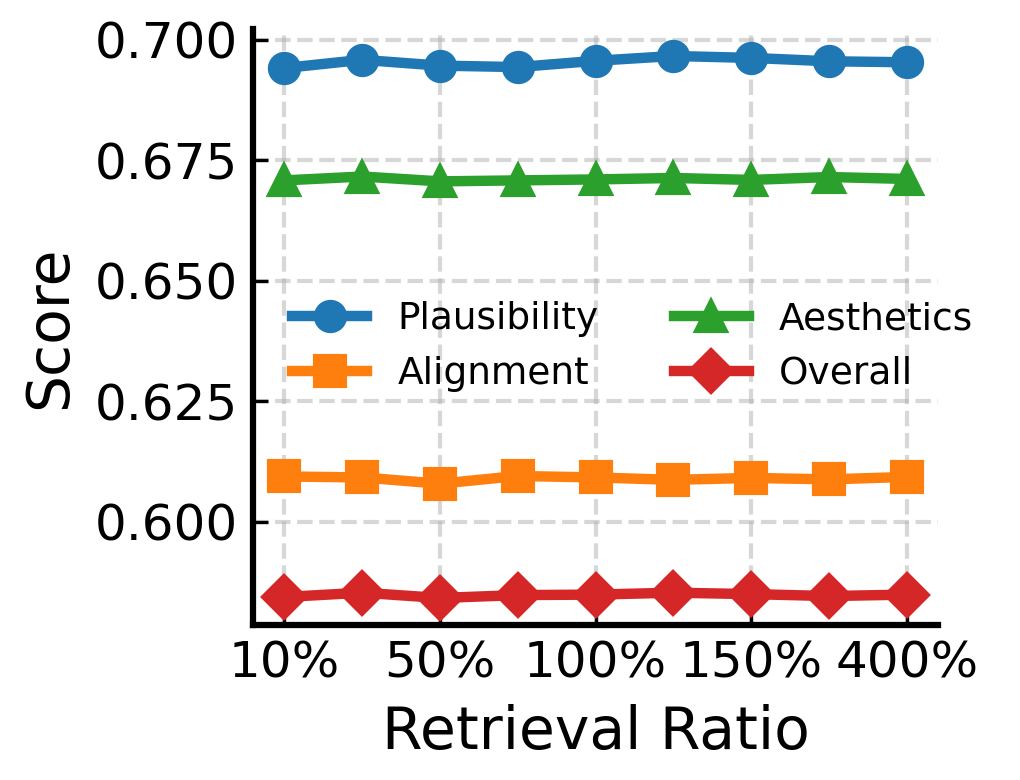} 
        \caption{RichHF-18K}
        \label{fig:ratio-richhf}
    \end{subfigure}

    \begin{subfigure}[b]{0.3\linewidth}
        \centering
        \includegraphics[width=\linewidth]{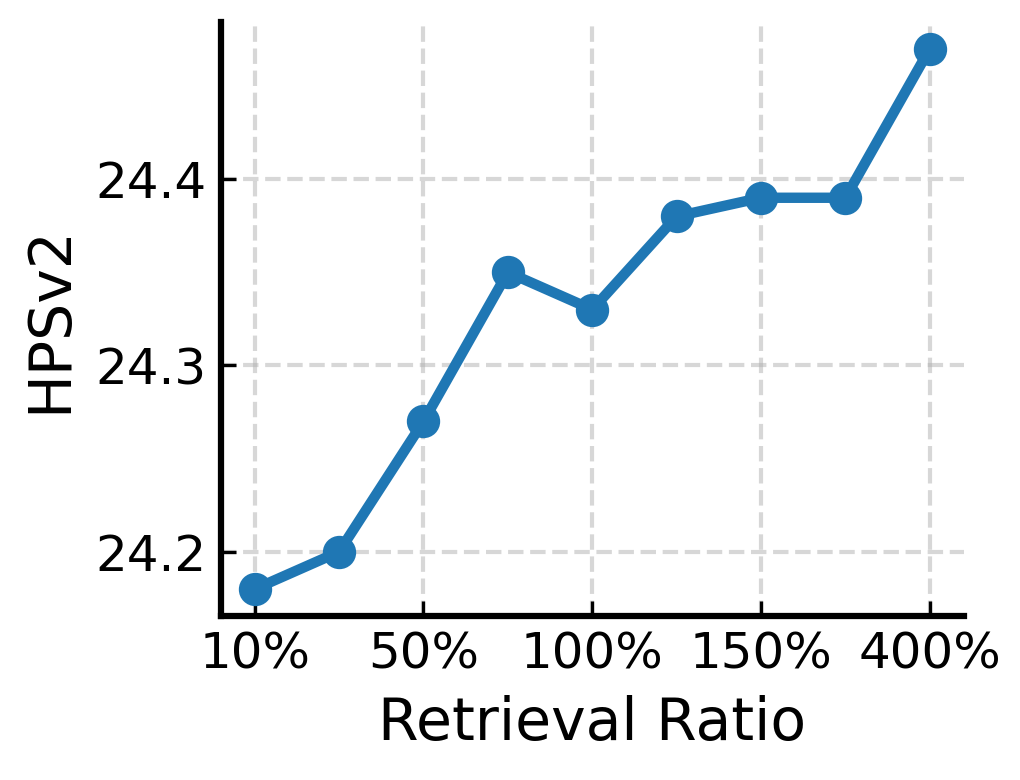}
        \caption{HPSv2}
        \label{fig:ratio-hps}
    \end{subfigure}
    \hfill
    \begin{subfigure}[b]{0.3\linewidth}
        \centering
        \includegraphics[width=\linewidth]{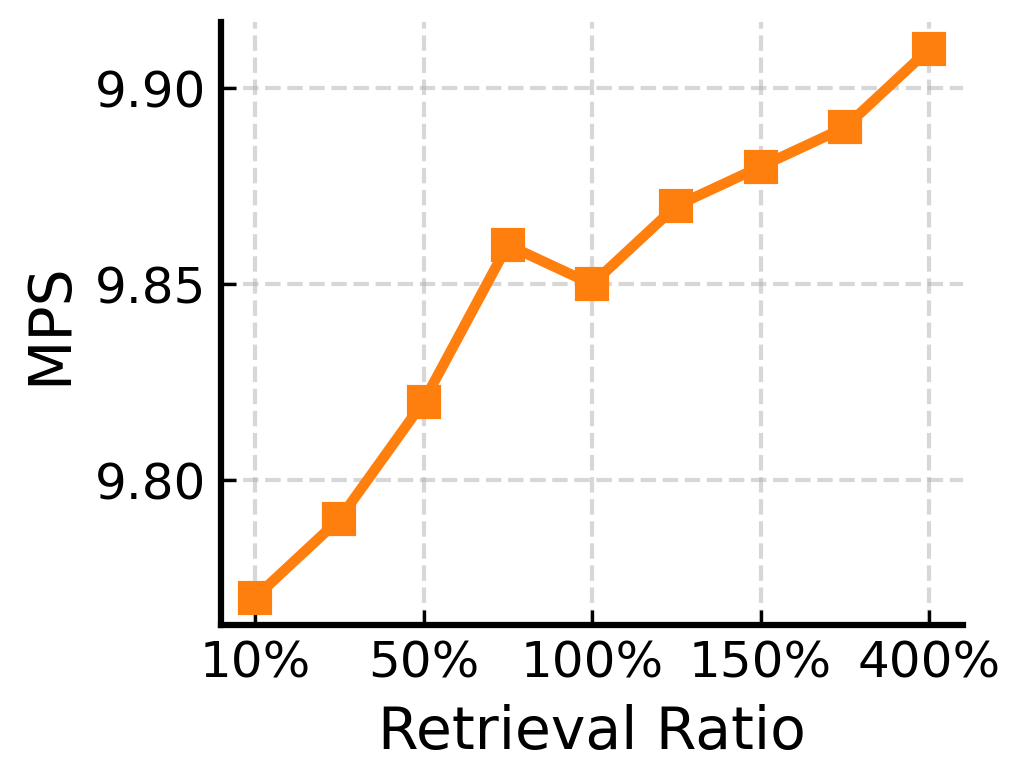}
        \caption{MPS}
        \label{fig:ratio-mps}
    \end{subfigure}
    \hfill
    \begin{subfigure}[b]{0.3\linewidth}
        \centering
        \includegraphics[width=\linewidth]{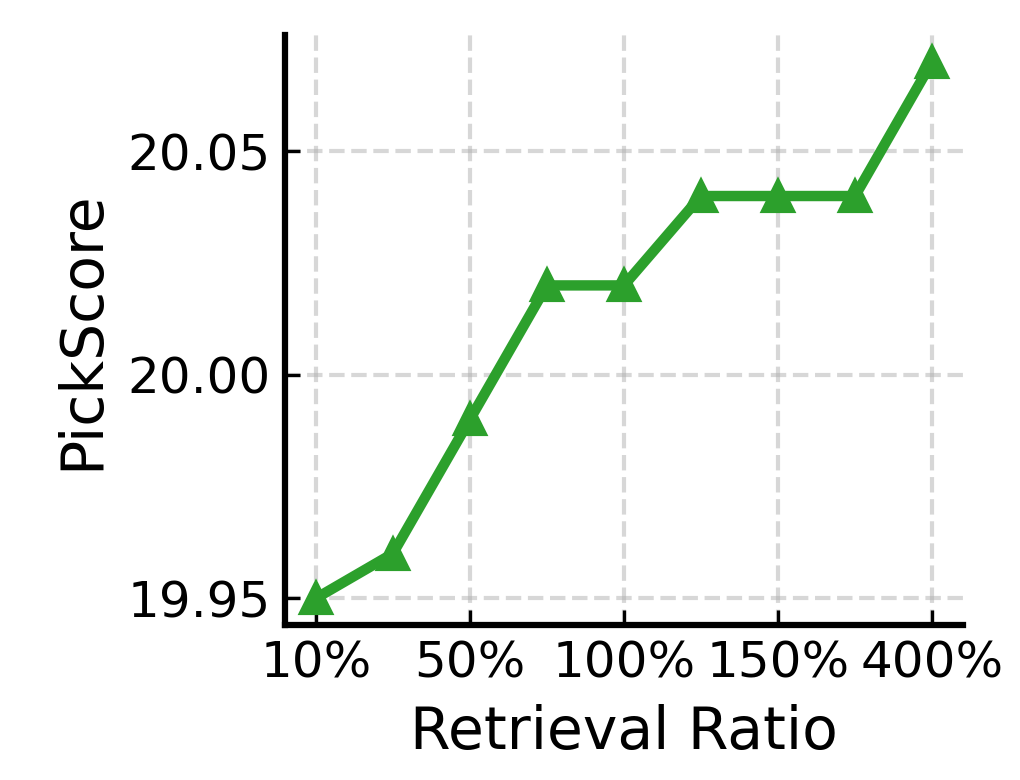}
        \caption{PickScore}
        \label{fig:ratio-pick}
    \end{subfigure}

    \caption{Ablation study on the retrieval ratio under DDIM sampler on a 5k subset of LAION.}
    \label{fig:Retrieval-Ratio}
\end{figure}

\noindent
\textbf{Rank-$k$ Retrieval Ablation.}
We ablate retrieval rank by replacing the highest-similarity image (Rank-1) with Rank-$k$ candidates in Tab. \ref{tab:component_ablation}.
As $k$ increases, text alignment declines due to lower prompt relevance, while realism metrics like Precision and Plausibility remain stable.
This suggests PixARG is robust to less relevant retrievals, as visual priors continue to support realistic synthesis.
% with retrieved priors continuing to support realistic synthesis.

\noindent
\textbf{Retrieval Ratio Ablation.}
We investigate how the retrieval database size influences performance by varying its ratio relative to the prompt counts.
As shown in Fig. \ref{fig:Retrieval-Ratio}, most metrics, including CLIP, FID, and human preference scores, improve steadily with a higher ratio.
Recall also rises, reflecting broader coverage and diversity.
This shows that a larger retrieval database offers more relevant and diverse visual exemplars, providing richer contextual priors for the diffusion process.
Meanwhile, realism metrics such as Precision and RichHF-18K plausibility remain stable across scales, suggesting that grounding on real visual priors inherently preserves realism and physical consistency regardless of database size.

\subsection{Online Retrieval Experiments}

\begin{figure}[!t]
    \centering
    \includegraphics[width=0.9\linewidth]{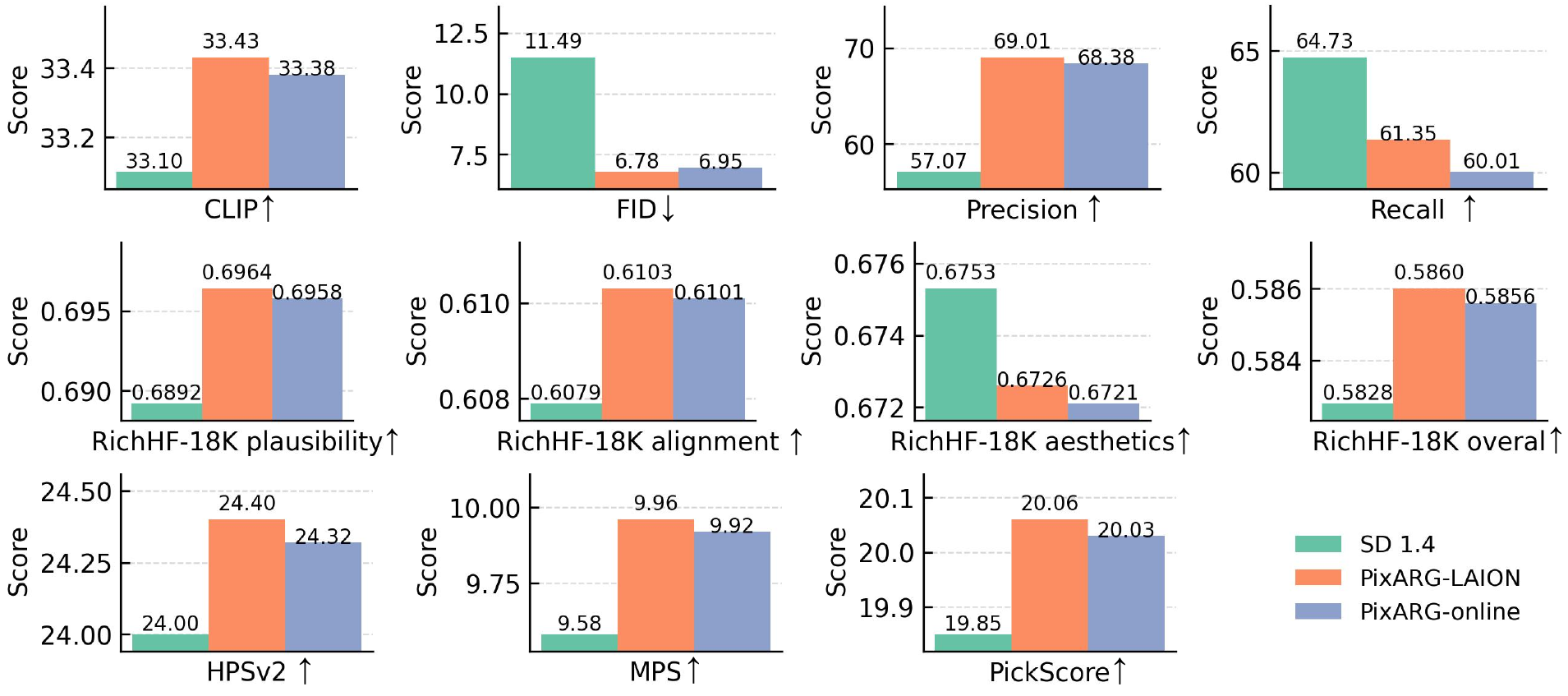}
    \caption{Online evaluation results using Google Image Search as the retrieval backend under the DDIM sampler.}
    \label{fig:online}
\end{figure}

To evaluate the generalization of our retrieval-augmented framework in real-world conditions, we conduct an online evaluation using Google Image Search as the retrieval engine, where top-ranked results serve as retrieved candidates.
As shown in Fig. \ref{fig:online}, PixARG consistently outperforms the SD 1.4 baseline across most metrics, demonstrating strong adaptability to open-domain retrieval sources.
Specifically, the CLIP score increases by 0.28 and Precision by 11.31\%.
When compared to the LAION-based retrieval setting, performance shows only a slight decline, mainly due to the uncurated nature of web retrieval, where low-quality or irrelevant images occasionally appear.
Despite this, the marginal drop indicates that PixARG maintains high robustness to retrieval noise and generalizes well to open-domain scenarios.

\subsection{User-Specific Retrieval Evaluation}
Unlike generic retrieval databases used in previous experiments, we evaluate performance in a user-specific setting where both prompts and retrieved images originate from the same user. Using RedCaps dataset, we conduct a per-user evaluation.
As detailed in Tab.~\ref{tab:user_performance}, PixARG consistently outperforms the SD 1.4 baseline across all users. We observe average improvements of 0.36 in CLIP alignment, 1.14\% in Recall, and significant gains in realism with 10.51\% higher Precision and 1.69\% higher Plausibility.
These results confirm that user-specific retrieval pools enable more realistic and aligned generation, as the higher consistency between user expressions and their own images ensures effective retrieval-based generation.

\begin{table}[!t]
\centering
\caption{Performance on user-specific datasets. $N$ denotes the total image-text pairs per user, where 20\% are used as query prompts and 80\% constitute the retrieval database.}
\renewcommand{\arraystretch}{0.8}
\resizebox{\linewidth}{!}{
\begin{tabular}{lllccccccc}
\toprule
\multirow{2}{*}{User Name} & \multirow{2}{*}{Pairs} & \multirow{2}{*}{Method} & \multicolumn{1}{c}{Basic} & \multicolumn{2}{c}{Realism \& Diversity} & \multicolumn{4}{c}{RichHF-18K} \\
\cmidrule(lr){4-4} \cmidrule(lr){5-6} \cmidrule(lr){7-10}
 & ($N$) & & CLIP↑ & Precision(\%)↑ & Reccall(\%)↑ & Plausibility↑ & Alignment↑ & Aesthetics↑ & Overall↑ \\ 
\midrule
& & SD 1.4 & 31.19 & 45.66 & 78.61 & 0.6227 & 0.5666 & \textbf{0.6001} & 0.5399 \\
\rowcolor{gray!10}\multirow{-2}{*}{earthmoonsun} & \multirow{-2}{*}{1736} & PixARG (Ours) & \textbf{31.52} & \textbf{58.96} & \textbf{81.50} & \textbf{0.6303} & \textbf{0.5676} & 0.5967 & \textbf{0.5411} \\
\midrule
& & SD 1.4 & 32.80 & 50.23 & 76.92 & 0.6480 & 0.5421 & 0.6200 & 0.5384 \\
\rowcolor{gray!10}\multirow{-2}{*}{ethan\_kahn} & \multirow{-2}{*}{1109} & PixARG (Ours) & \textbf{33.01} & \textbf{70.59} & \textbf{77.38} & \textbf{0.6609} & \textbf{0.5435} & \textbf{0.6305} & \textbf{0.5431} \\
\midrule
& & SD 1.4 & 28.81 & 62.50 & 71.88 & 0.6429 & 0.6177 & \textbf{0.6282} & 0.5621 \\
\rowcolor{gray!10}\multirow{-2}{*}{GallowBoob} & \multirow{-2}{*}{1285} & PixARG (Ours) & \textbf{29.33} & \textbf{72.66} & \textbf{73.44} & \textbf{0.6504} & \textbf{0.6200} & 0.6166 & \textbf{0.5665} \\
\midrule
& & SD 1.4 & 30.99 & 55.23 & \textbf{83.04} & 0.7703 & 0.4922 & 0.6716 & 0.5531 \\
\rowcolor{gray!10}\multirow{-2}{*}{Masslm01973} & \multirow{-2}{*}{1387} & PixARG (Ours) & \textbf{31.20} & \textbf{69.68} & 83.03 & \textbf{0.7738} & \textbf{0.4929} & \textbf{0.6772} & \textbf{0.5541} \\
\midrule
& & SD 1.4 & 29.77 & 73.33 & 84.29 & 0.6500 & 0.6079 & 0.6299 & 0.5659 \\
\rowcolor{gray!10}\multirow{-2}{*}{mike\_pants} & \multirow{-2}{*}{1052} & PixARG (Ours) & \textbf{30.39} & \textbf{73.38} & \textbf{85.71} & \textbf{0.6835} & \textbf{0.6216} & \textbf{0.6418} & \textbf{0.5838} \\
\midrule
& & SD 1.4 & 31.53 & 74.64 & 81.30 & 0.7538 & 0.6884 & \textbf{0.7115} & 0.6497 \\
\rowcolor{gray!10}\multirow{-2}{*}{Proteon} & \multirow{-2}{*}{2090} & PixARG (Ours) & \textbf{31.80} & \textbf{79.43} & \textbf{81.82} & \textbf{0.7550} & \textbf{0.6886} & 0.7077 & \textbf{0.6516} \\
\bottomrule
\end{tabular}}
\label{tab:user_performance}
\end{table}

\subsection{Experiments on More Diffusion Backbones}

\begin{table}[t]
\centering
\caption{Evaluation on other diffusion backbones. “+ Ours” indicates the integration of PixARG.}
\renewcommand{\arraystretch}{0.8}
\resizebox{\linewidth}{!}{
\begin{tabular}{lcccccc}
\toprule
\multirow{2}{*}{Backbone} & \multirow{2}{*}{Steps} & 
\multicolumn{2}{c}{Basic Metrics} & 
\multicolumn{2}{c}{Realism \& Diversity} & 
\multicolumn{1}{c}{RichHF-18K} \\
\cmidrule(lr){3-4} \cmidrule(lr){5-6} \cmidrule(lr){7-7}
& & CLIP↑ & FID↓ & Precision(\%)↑ & Recall(\%)↑ & Overall↑ \\ 
\midrule
PixArt-$\alpha$~\cite{chen2023pixart} & 15 & 31.55 & 18.75 & 48.16 & 46.73 & \textbf{0.6183} \\
\rowcolor{gray!10}+ Ours & \textbf{8} & \textbf{31.63} & \textbf{12.58} & \textbf{57.86} & \textbf{54.72} & 0.6037 \\
\midrule
SDXL~\cite{podell2023sdxl} & 50 & 32.24 & 14.22 & 65.33 & 62.75 & 0.5248 \\
\rowcolor{gray!10}+ Ours & \textbf{15} & \textbf{32.32} & \textbf{7.31} & \textbf{68.29} & \textbf{68.70} & \textbf{0.5287} \\
\bottomrule
\end{tabular}}
\label{tab:other_models}
\end{table}

We further evaluate PixARG on two representative backbones, PixArt-$\alpha$ and SDXL, to test generalization.
As shown in Tab. \ref{tab:other_models}, PixARG delivers comparable or superior performance while requiring only 53\% and 30\% diffusion steps. 
PixARG improves Precision by +9.7\% and +2.96\%, respectively, demonstrating the effectiveness of incorporating real-image priors in enhancing fidelity.
Meanwhile, FID decreases by 6.17–6.91, and Recall increases by 7.99–5.95\%, indicating that the retrieval database enhances generation diversity without sacrificing fidelity.
These results demonstrate that our retrieval method's gains in enhancing visual realism and accelerating generation are significant and broadly applicable.

\subsection{Visualization}

\begin{figure}[!t]
    \centering
    \includegraphics[width=0.9\linewidth]{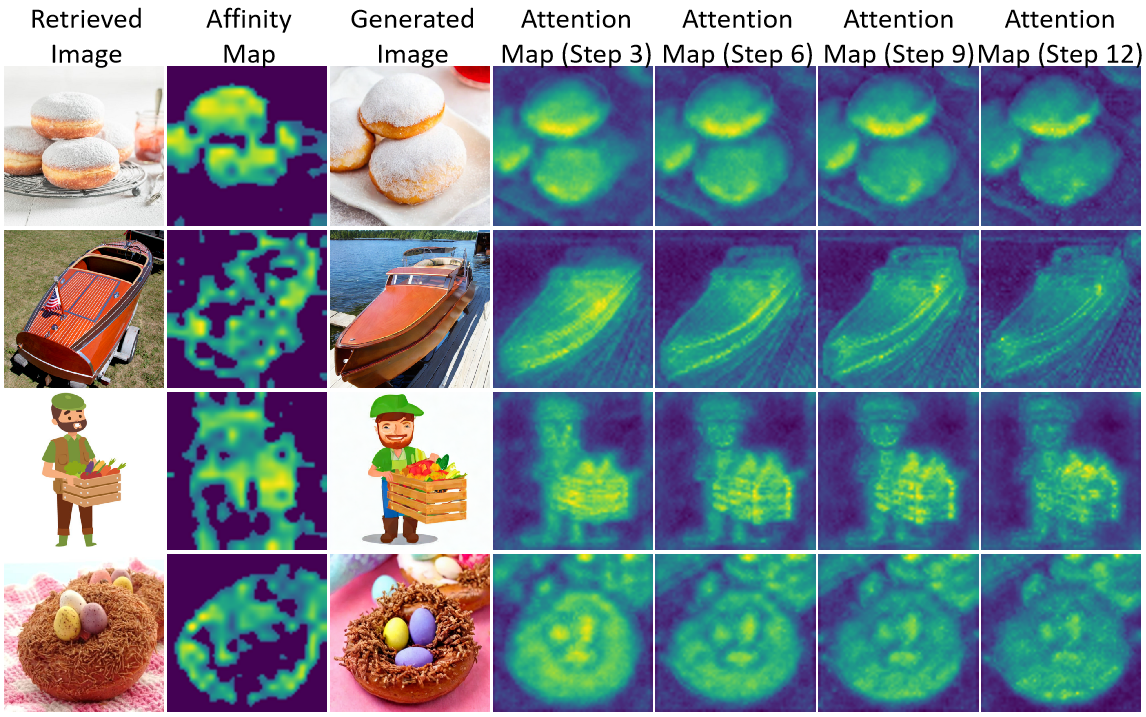}
    \caption{Visualization of retrieval items and attention maps on LAION dataset~\cite{schuhmann2022laion}. The text-visual affinity maps are highly consistent with the model’s attention maps, providing effective visual prior for efficient generation.}
    \label{fig:visual-attention}
\end{figure}

We visualize the retrieved images, text-visual affinity maps, generated outputs, and attention maps in Fig. \ref{fig:visual-attention}.
The affinity maps are closely match the model’s attention, indicating retrieval provides effective visual priors.
High-alignment regions retain richer structural details and receive stronger attention.
In our framework, these regions are perturbed with weaker noise and denoised more gradually, enabling fine-grained reconstruction even with fewer diffusion steps.
Conversely, low-alignment regions, typically background or semantically irrelevant areas, receive stronger noise in the forward process and are denoised more aggressively. These regions exhibit weak attention and contribute little to the overall semantics, so their minor quality degradation is acceptable in exchange for faster convergence.
This adaptive design allows PixARG to accelerate generation while preserving the fidelity of key regions.

\begin{figure}[!t]
    \centering
    \includegraphics[width=0.9\linewidth]{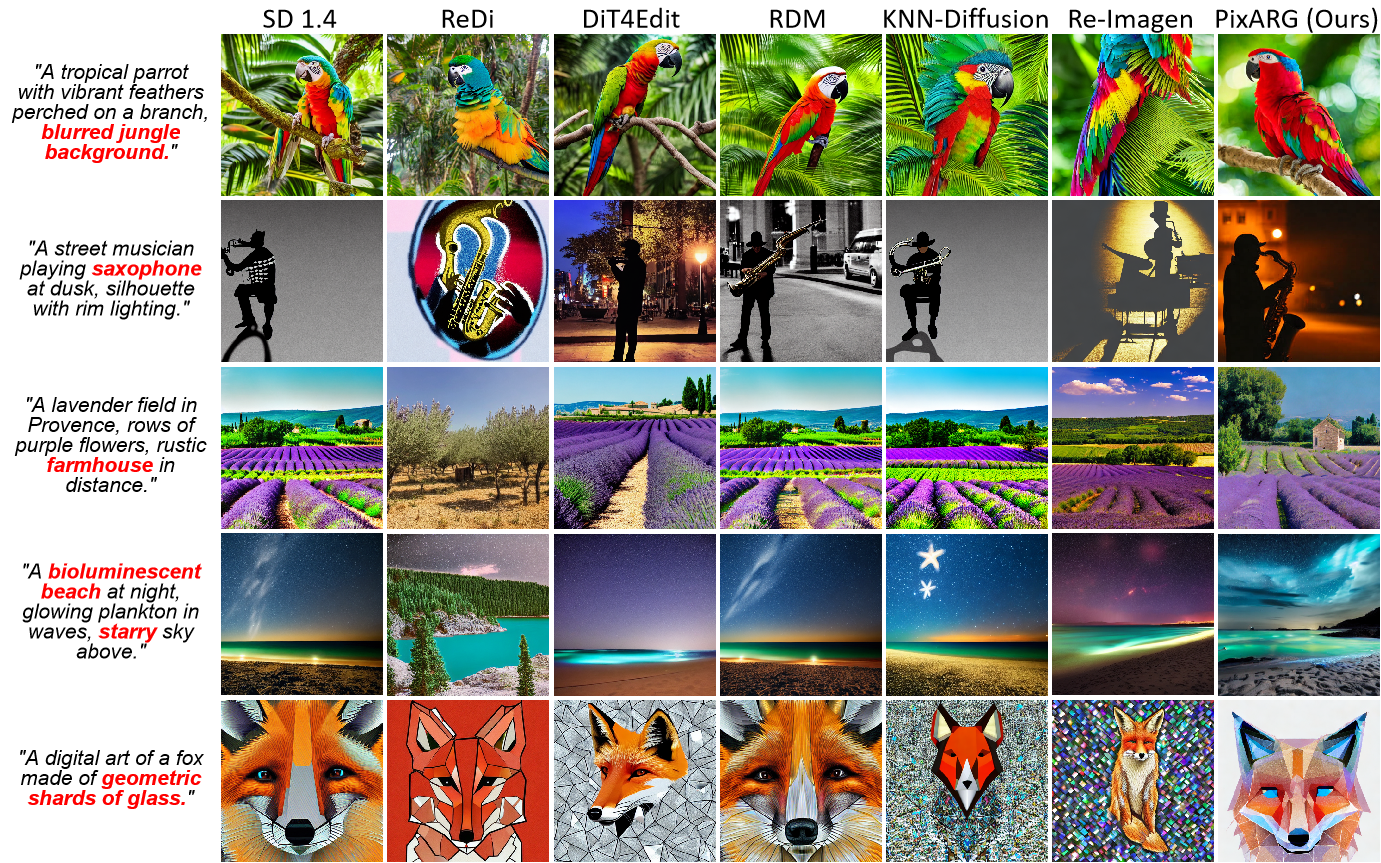}
    \caption{Visual comparison with other methods under the same settings as Tab. \ref{tab:main_results}.}
    \label{fig:visual-compare}
\end{figure}

Fig. \ref{fig:visual-compare} presents a qualitative comparison between our method and existing baselines. Notably, these baselines fail to demonstrate significant improvement over the standard SD baseline. This limitation stems primarily from the fact that they are either not tailored for retrieval-based generation or focus exclusively on generating unseen objects, thus preventing them from leveraging the rich visual priors and fine-grained textures inherent in real retrieved images. In contrast, by effectively integrating these retrieved features, our model consistently generates images with superior visual quality, fidelity, and realism across diverse prompts.

\section{Conclusion}

In this work, we introduce pixel-asynchronous retrieval-augmented generation (PixARG), a novel pixel-level retrieval-augmented diffusion framework that integrates real images to enable fast and high-fidelity text-to-image generation. PixARG employs a pixel-level text-visual affinity map to identify concept-relevant pixel information and injects spatially heterogeneous noise to selectively preserve informative regions. A pixel-wise asynchronous denoising is then applied to assign denoising rates based on noise intensity, while conditioning the model on the affinity map with a tailored loss to handle heterogeneous noise. Benefiting from an information-enriched initialization, PixARG achieves fast and realistic generation. Extensive experiments validate the effectiveness of PixARG across different retrieval settings and diffusion backbones.

%%
%% The acknowledgments section is defined using the "acks" environment
%% (and NOT an unnumbered section). This ensures the proper
%% identification of the section in the article metadata, and the
%% consistent spelling of the heading.
\begin{acks}
This work was supported in part by China NSF grant (No. 62572299, No. 62202297, No. 62441236, No. 62372296, No. 62432007, No. U24A20326, No. U25A6024, No. U25A20437), Alibaba Innovation Research (AIR) Program, Tencent WeChat Research Program, and SJTU-Huawei Explore X Gift Fund.
\end{acks}

% \section*{Ethics and Privacy Statement}

% This section of your ACM work should discuss the potential societal
% risks that might result from its publication; two to three sentences
% related to the findings of your study, or new advancements made
% possible by their developed methods. The privacy and ethics statement
% should clearly address the broader impacts of their work as it relates
% to the authors' interpretation of privacy, fairness, safety, human
% rights, data sovereignty, or future misuse and any benefit/risk
% trade-off resulting from this research. We acknowledge that some
% papers may have minimal societal risks beyond those considered by
% institutional review boards, and the dimensions considered by any
% review of the user study design or dataset licenses could be provided
% in this statement.

%%
%% The next two lines define the bibliography style to be used, and
%% the bibliography file.
\bibliographystyle{ACM-Reference-Format}
\balance
\bibliography{sample-base}

@String{BIT = "{BIT}" }

@String{Computer = "{IEEE} Computer" }

@String{Springer = "Springer-Verlag" }

@inproceedings{li2023q,
  title={Q-diffusion: Quantizing diffusion models},
  author={Li, Xiuyu and Liu, Yijiang and Lian, Long and Yang, Huanrui and Dong, Zhen and Kang, Daniel and Zhang, Shanghang and Keutzer, Kurt},
  booktitle={Proceedings of the IEEE/CVF International Conference on Computer Vision},
  pages={17535--17545},
  year={2023}
}

@article{li2024svdquant,
  title={Svdquant: Absorbing outliers by low-rank components for 4-bit diffusion models},
  author={Li, Muyang and Lin, Yujun and Zhang, Zhekai and Cai, Tianle and Li, Xiuyu and Guo, Junxian and Xie, Enze and Meng, Chenlin and Zhu, Jun-Yan and Han, Song},
  journal={arXiv preprint arXiv:2411.05007},
  year={2024}
}

@article{blattmann2022retrieval,
  title={Retrieval-augmented diffusion models},
  author={Blattmann, Andreas and Rombach, Robin and Oktay, Kaan and M{\"u}ller, Jonas and Ommer, Bj{\"o}rn},
  journal={Advances in Neural Information Processing Systems},
  volume={35},
  pages={15309--15324},
  year={2022}
}

@article{sheynin2022knn,
  title={Knn-diffusion: Image generation via large-scale retrieval},
  author={Sheynin, Shelly and Ashual, Oron and Polyak, Adam and Singer, Uriel and Gafni, Oran and Nachmani, Eliya and Taigman, Yaniv},
  journal={arXiv preprint arXiv:2204.02849},
  year={2022}
}

@article{chen2022re,
  title={Re-imagen: Retrieval-augmented text-to-image generator},
  author={Chen, Wenhu and Hu, Hexiang and Saharia, Chitwan and Cohen, William W},
  journal={arXiv preprint arXiv:2209.14491},
  year={2022}
}

@inproceedings{zhang2023redi,
  title={ReDi: efficient learning-free diffusion inference via trajectory retrieval},
  author={Zhang, Kexun and Yang, Xianjun and Wang, William Yang and Li, Lei},
  booktitle={International Conference on Machine Learning},
  pages={41770--41785},
  year={2023},
  organization={PMLR}
}

@misc{kakaobrain2022coyo-700m,
  title         = {COYO-700M: Image-Text Pair Dataset},
  author        = {Byeon, Minwoo and Park, Beomhee and Kim, Haecheon and Lee, Sungjun and Baek, Woonhyuk and Kim, Saehoon},
  year          = {2022},
  howpublished  = {\url{https://github.com/kakaobrain/coyo-dataset}},
}

@inproceedings{sdv1,
  title={High-resolution image synthesis with latent diffusion models},
  author={Rombach, Robin and Blattmann, Andreas and Lorenz, Dominik and Esser, Patrick and Ommer, Bj{\"o}rn},
  booktitle={Proceedings of the IEEE/CVF conference on computer vision and pattern recognition},
  pages={10684--10695},
  year={2022}
}

@inproceedings{sd3,
  title={Scaling rectified flow transformers for high-resolution image synthesis},
  author={Esser, Patrick and Kulal, Sumith and Blattmann, Andreas and Entezari, Rahim and M{\"u}ller, Jonas and Saini, Harry and Levi, Yam and Lorenz, Dominik and Sauer, Axel and Boesel, Frederic and others},
  booktitle={Forty-first international conference on machine learning},
  year={2024}
}

@article{imagen3,
  title={Imagen 3},
  author={Baldridge, Jason and Bauer, Jakob and Bhutani, Mukul and Brichtova, Nicole and Bunner, Andrew and Castrejon, Lluis and Chan, Kelvin and Chen, Yichang and Dieleman, Sander and Du, Yuqing and others},
  journal={arXiv preprint arXiv:2408.07009},
  year={2024}
}

@article{FLUX.1Kontext,
  title={FLUX. 1 Kontext: Flow Matching for In-Context Image Generation and Editing in Latent Space},
  author={Batifol, Stephen and Blattmann, Andreas and Boesel, Frederic and Consul, Saksham and Diagne, Cyril and Dockhorn, Tim and English, Jack and English, Zion and Esser, Patrick and Kulal, Sumith and others},
  journal={arXiv e-prints},
  pages={arXiv--2506},
  year={2025}
}

@article{dalle3,
  title={Improving image generation with better captions},
  author={Betker, James and Goh, Gabriel and Jing, Li and Brooks, Tim and Wang, Jianfeng and Li, Linjie and Ouyang, Long and Zhuang, Juntang and Lee, Joyce and Guo, Yufei and others},
  journal={Computer Science. https://cdn. openai. com/papers/dall-e-3. pdf},
  volume={2},
  number={3},
  pages={8},
  year={2023}
}

@article{zhang2024laptop,
  title={Laptop-diff: Layer pruning and normalized distillation for compressing diffusion models},
  author={Zhang, Dingkun and Li, Sijia and Chen, Chen and Xie, Qingsong and Lu, Haonan},
  journal={arXiv preprint arXiv:2404.11098},
  year={2024}
}

@inproceedings{shen2025lazydit,
  title={Lazydit: Lazy learning for the acceleration of diffusion transformers},
  author={Shen, Xuan and Song, Zhao and Zhou, Yufa and Chen, Bo and Li, Yanyu and Gong, Yifan and Zhang, Kai and Tan, Hao and Kuen, Jason and Ding, Henghui and others},
  booktitle={Proceedings of the AAAI Conference on Artificial Intelligence},
  volume={39},
  number={19},
  pages={20409--20417},
  year={2025}
}

@inproceedings{kahatapitiya2025adaptive,
  title={Adaptive caching for faster video generation with diffusion transformers},
  author={Kahatapitiya, Kumara and Liu, Haozhe and He, Sen and Liu, Ding and Jia, Menglin and Zhang, Chenyang and Ryoo, Michael S and Xie, Tian},
  booktitle={Proceedings of the IEEE/CVF International Conference on Computer Vision},
  pages={15240--15252},
  year={2025}
}

@inproceedings{yin2024one,
  title={One-step diffusion with distribution matching distillation},
  author={Yin, Tianwei and Gharbi, Micha{\"e}l and Zhang, Richard and Shechtman, Eli and Durand, Fredo and Freeman, William T and Park, Taesung},
  booktitle={Proceedings of the IEEE/CVF conference on computer vision and pattern recognition},
  pages={6613--6623},
  year={2024}
}

@article{luo2024one,
  title={One-step diffusion distillation through score implicit matching},
  author={Luo, Weijian and Huang, Zemin and Geng, Zhengyang and Kolter, J Zico and Qi, Guo-jun},
  journal={Advances in Neural Information Processing Systems},
  volume={37},
  pages={115377--115408},
  year={2024}
}

@article{gao2023retrieval,
  title={Retrieval-augmented generation for large language models: A survey},
  author={Gao, Yunfan and Xiong, Yun and Gao, Xinyu and Jia, Kangxiang and Pan, Jinliu and Bi, Yuxi and Dai, Yixin and Sun, Jiawei and Wang, Haofen and Wang, Haofen},
  journal={arXiv preprint arXiv:2312.10997},
  volume={2},
  number={1},
  year={2023}
}

@article{lewis2020retrieval,
  title={Retrieval-augmented generation for knowledge-intensive nlp tasks},
  author={Lewis, Patrick and Perez, Ethan and Piktus, Aleksandra and Petroni, Fabio and Karpukhin, Vladimir and Goyal, Naman and K{\"u}ttler, Heinrich and Lewis, Mike and Yih, Wen-tau and Rockt{\"a}schel, Tim and others},
  journal={Advances in neural information processing systems},
  volume={33},
  pages={9459--9474},
  year={2020}
}

@article{guan2025deeprag,
  title={DeepRAG: Thinking to Retrieve Step by Step for Large Language Models},
  author={Guan, Xinyan and Zeng, Jiali and Meng, Fandong and Xin, Chunlei and Lu, Yaojie and Lin, Hongyu and Han, Xianpei and Sun, Le and Zhou, Jie},
  journal={arXiv preprint arXiv:2502.01142},
  year={2025}
}

@article{izacard2023atlas,
  title={Atlas: Few-shot learning with retrieval augmented language models},
  author={Izacard, Gautier and Lewis, Patrick and Lomeli, Maria and Hosseini, Lucas and Petroni, Fabio and Schick, Timo and Dwivedi-Yu, Jane and Joulin, Armand and Riedel, Sebastian and Grave, Edouard},
  journal={Journal of Machine Learning Research},
  volume={24},
  number={251},
  pages={1--43},
  year={2023}
}

@article{huang2024less,
  title={Less is more: Making smaller language models competent subgraph retrievers for multi-hop kgqa},
  author={Huang, Wenyu and Zhou, Guancheng and Wang, Hongru and Vougiouklis, Pavlos and Lapata, Mirella and Pan, Jeff Z},
  journal={arXiv preprint arXiv:2410.06121},
  year={2024}
}

@article{kashmira2024tobugraph,
  title={TOBUGraph: Knowledge Graph-Based Retrieval for Enhanced LLM Performance Beyond RAG},
  author={Kashmira, Savini and Dantanarayana, Jayanaka L and Brodsky, Joshua and Mahendra, Ashish and Kang, Yiping and Flautner, Krisztian and Tang, Lingjia and Mars, Jason},
  journal={arXiv preprint arXiv:2412.05447},
  year={2024}
}

@article{schuhmann2022laion,
  title={Laion-5b: An open large-scale dataset for training next generation image-text models},
  author={Schuhmann, Christoph and Beaumont, Romain and Vencu, Richard and Gordon, Cade and Wightman, Ross and Cherti, Mehdi and Coombes, Theo and Katta, Aarush and Mullis, Clayton and Wortsman, Mitchell and others},
  journal={Advances in neural information processing systems},
  volume={35},
  pages={25278--25294},
  year={2022}
}

@article{chen2023pixart,
  title={Pixart-{$\alpha$}: Fast training of diffusion transformer for photorealistic text-to-image synthesis},
  author={Chen, Junsong and Yu, Jincheng and Ge, Chongjian and Yao, Lewei and Xie, Enze and Wu, Yue and Wang, Zhongdao and Kwok, James and Luo, Ping and Lu, Huchuan and others},
  journal={arXiv preprint arXiv:2310.00426},
  year={2023}
}

@article{podell2023sdxl,
  title={Sdxl: Improving latent diffusion models for high-resolution image synthesis},
  author={Podell, Dustin and English, Zion and Lacey, Kyle and Blattmann, Andreas and Dockhorn, Tim and M{\"u}ller, Jonas and Penna, Joe and Rombach, Robin},
  journal={arXiv preprint arXiv:2307.01952},
  year={2023}
}

@article{song2020denoising,
  title={Denoising diffusion implicit models},
  author={Song, Jiaming and Meng, Chenlin and Ermon, Stefano},
  journal={arXiv preprint arXiv:2010.02502},
  year={2020}
}

@article{liu2022pseudo,
  title={Pseudo numerical methods for diffusion models on manifolds},
  author={Liu, Luping and Ren, Yi and Lin, Zhijie and Zhao, Zhou},
  journal={arXiv preprint arXiv:2202.09778},
  year={2022}
}

@article{lu2025dpm,
  title={Dpm-solver++: Fast solver for guided sampling of diffusion probabilistic models},
  author={Lu, Cheng and Zhou, Yuhao and Bao, Fan and Chen, Jianfei and Li, Chongxuan and Zhu, Jun},
  journal={Machine Intelligence Research},
  pages={1--22},
  year={2025},
  publisher={Springer}
}

@article{karras2022elucidating,
  title={Elucidating the design space of diffusion-based generative models},
  author={Karras, Tero and Aittala, Miika and Aila, Timo and Laine, Samuli},
  journal={Advances in neural information processing systems},
  volume={35},
  pages={26565--26577},
  year={2022}
}

@article{gandelsman2023interpreting,
  title={Interpreting clip's image representation via text-based decomposition},
  author={Gandelsman, Yossi and Efros, Alexei A and Steinhardt, Jacob},
  journal={arXiv preprint arXiv:2310.05916},
  year={2023}
}

@inproceedings{feng2025dit4edit,
  title={Dit4edit: Diffusion transformer for image editing},
  author={Feng, Kunyu and Ma, Yue and Wang, Bingyuan and Qi, Chenyang and Chen, Haozhe and Chen, Qifeng and Wang, Zeyu},
  booktitle={Proceedings of the AAAI Conference on Artificial Intelligence},
  volume={39},
  number={3},
  pages={2969--2977},
  year={2025}
}

@article{hu2022lora,
  title={Lora: Low-rank adaptation of large language models.},
  author={Hu, Edward J and Shen, Yelong and Wallis, Phillip and Allen-Zhu, Zeyuan and Li, Yuanzhi and Wang, Shean and Wang, Lu and Chen, Weizhu and others},
  journal={ICLR},
  volume={1},
  number={2},
  pages={3},
  year={2022}
}

@inproceedings{radford2021learning,
  title={Learning transferable visual models from natural language supervision},
  author={Radford, Alec and Kim, Jong Wook and Hallacy, Chris and Ramesh, Aditya and Goh, Gabriel and Agarwal, Sandhini and Sastry, Girish and Askell, Amanda and Mishkin, Pamela and Clark, Jack and others},
  booktitle={International conference on machine learning},
  pages={8748--8763},
  year={2021},
  organization={PmLR}
}

@article{heusel2017gans,
  title={Gans trained by a two time-scale update rule converge to a local nash equilibrium},
  author={Heusel, Martin and Ramsauer, Hubert and Unterthiner, Thomas and Nessler, Bernhard and Hochreiter, Sepp},
  journal={Advances in neural information processing systems},
  volume={30},
  year={2017}
}

@article{kynkaanniemi2019improved,
  title={Improved precision and recall metric for assessing generative models},
  author={Kynk{\"a}{\"a}nniemi, Tuomas and Karras, Tero and Laine, Samuli and Lehtinen, Jaakko and Aila, Timo},
  journal={Advances in neural information processing systems},
  volume={32},
  year={2019}
}

@inproceedings{liang2024rich,
  title={Rich human feedback for text-to-image generation},
  author={Liang, Youwei and He, Junfeng and Li, Gang and Li, Peizhao and Klimovskiy, Arseniy and Carolan, Nicholas and Sun, Jiao and Pont-Tuset, Jordi and Young, Sarah and Yang, Feng and others},
  booktitle={Proceedings of the IEEE/CVF Conference on Computer Vision and Pattern Recognition},
  pages={19401--19411},
  year={2024}
}

@article{wu2023human,
  title={Human preference score v2: A solid benchmark for evaluating human preferences of text-to-image synthesis},
  author={Wu, Xiaoshi and Hao, Yiming and Sun, Keqiang and Chen, Yixiong and Zhu, Feng and Zhao, Rui and Li, Hongsheng},
  journal={arXiv preprint arXiv:2306.09341},
  year={2023}
}

@inproceedings{zhang2024learning,
  title={Learning multi-dimensional human preference for text-to-image generation},
  author={Zhang, Sixian and Wang, Bohan and Wu, Junqiang and Li, Yan and Gao, Tingting and Zhang, Di and Wang, Zhongyuan},
  booktitle={Proceedings of the IEEE/CVF Conference on Computer Vision and Pattern Recognition},
  pages={8018--8027},
  year={2024}
}

@article{kirstain2023pick,
  title={Pick-a-pic: An open dataset of user preferences for text-to-image generation},
  author={Kirstain, Yuval and Polyak, Adam and Singer, Uriel and Matiana, Shahbuland and Penna, Joe and Levy, Omer},
  journal={Advances in neural information processing systems},
  volume={36},
  pages={36652--36663},
  year={2023}
}

@article{ho2020denoising,
  title={Denoising diffusion probabilistic models},
  author={Ho, Jonathan and Jain, Ajay and Abbeel, Pieter},
  journal={Advances in neural information processing systems},
  volume={33},
  pages={6840--6851},
  year={2020}
}

@article{yan2024hybrid,
  title={Hybrid sd: Edge-cloud collaborative inference for stable diffusion models},
  author={Yan, Chenqian and Liu, Songwei and Liu, Hongjian and Peng, Xurui and Wang, Xiaojian and Chen, Fangmin and Fu, Lean and Mei, Xing},
  journal={arXiv preprint arXiv:2408.06646},
  year={2024}
}

@inproceedings{fang2025tinyfusion,
  title={Tinyfusion: Diffusion transformers learned shallow},
  author={Fang, Gongfan and Li, Kunjun and Ma, Xinyin and Wang, Xinchao},
  booktitle={Proceedings of the Computer Vision and Pattern Recognition Conference},
  pages={18144--18154},
  year={2025}
}

@inproceedings{liu2023instaflow,
  title={Instaflow: One step is enough for high-quality diffusion-based text-to-image generation},
  author={Liu, Xingchao and Zhang, Xiwen and Ma, Jianzhu and Peng, Jian and others},
  booktitle={The Twelfth International Conference on Learning Representations},
  year={2023}
}

@inproceedings{ma2024deepcache,
  title={Deepcache: Accelerating diffusion models for free},
  author={Ma, Xinyin and Fang, Gongfan and Wang, Xinchao},
  booktitle={Proceedings of the IEEE/CVF conference on computer vision and pattern recognition},
  pages={15762--15772},
  year={2024}
}

@article{liu2025reusing,
  title={From reusing to forecasting: Accelerating diffusion models with taylorseers},
  author={Liu, Jiacheng and Zou, Chang and Lyu, Yuanhuiyi and Chen, Junjie and Zhang, Linfeng},
  journal={arXiv preprint arXiv:2503.06923},
  year={2025}
}

@inproceedings{ye2025schedule,
  title={Schedule on the fly: Diffusion time prediction for faster and better image generation},
  author={Ye, Zilyu and Chen, Zhiyang and Li, Tiancheng and Huang, Zemin and Luo, Weijian and Qi, Guo-Jun},
  booktitle={Proceedings of the Computer Vision and Pattern Recognition Conference},
  pages={23412--23422},
  year={2025}
}

@article{huang2025illume+,
  title={Illume+: Illuminating unified mllm with dual visual tokenization and diffusion refinement},
  author={Huang, Runhui and Wang, Chunwei and Yang, Junwei and Lu, Guansong and Yuan, Yunlong and Han, Jianhua and Hou, Lu and Zhang, Wei and Hong, Lanqing and Zhao, Hengshuang and others},
  journal={arXiv preprint arXiv:2504.01934},
  year={2025}
}

@article{luo2023latent,
  title={Latent consistency models: Synthesizing high-resolution images with few-step inference},
  author={Luo, Simian and Tan, Yiqin and Huang, Longbo and Li, Jian and Zhao, Hang},
  journal={arXiv preprint arXiv:2310.04378},
  year={2023}
}

@article{sun2025noise,
  title={Is Noise Conditioning Necessary for Denoising Generative Models?},
  author={Sun, Qiao and Jiang, Zhicheng and Zhao, Hanhong and He, Kaiming},
  journal={arXiv preprint arXiv:2502.13129},
  year={2025}
}

@article{desai2021redcaps,
  title={Redcaps: Web-curated image-text data created by the people, for the people},
  author={Desai, Karan and Kaul, Gaurav and Aysola, Zubin and Johnson, Justin},
  journal={arXiv preprint arXiv:2111.11431},
  year={2021}
}

@inproceedings{wang2024crafting,
  title={Crafting personalized agents through retrieval-augmented generation on editable memory graphs},
  author={Wang, Zheng and Li, Zhongyang and Jiang, Zeren and Tu, Dandan and Shi, Wei},
  booktitle={Proceedings of the 2024 Conference on Empirical Methods in Natural Language Processing},
  pages={4891--4906},
  year={2024}
}

@article{salemi2026improving,
  title={Improving User Privacy in Personalized Generation: Client-Side Retrieval-Augmented Modification of Server-Side Generated Speculations},
  author={Salemi, Alireza and Zamani, Hamed},
  journal={arXiv preprint arXiv:2601.17569},
  year={2026}
}

@inproceedings{wang2024mememo,
  title={Mememo: On-device retrieval augmentation for private and personalized text generation},
  author={Wang, Zijie J and Chau, Duen Horng},
  booktitle={Proceedings of the 47th International ACM SIGIR Conference on Research and Development in Information Retrieval},
  pages={2765--2770},
  year={2024}
}

@inproceedings{lugmayr2022repaint,
  title={Repaint: Inpainting using denoising diffusion probabilistic models},
  author={Lugmayr, Andreas and Danelljan, Martin and Romero, Andres and Yu, Fisher and Timofte, Radu and Van Gool, Luc},
  booktitle={2022 IEEE/CVF conference on computer vision and pattern recognition (CVPR)},
  pages={11451--11461},
  year={2022},
  organization={IEEE}
}

@inproceedings{brack2024ledits++,
  title={Ledits++: Limitless image editing using text-to-image models},
  author={Brack, Manuel and Friedrich, Felix and Kornmeier, Katharia and Tsaban, Linoy and Schramowski, Patrick and Kersting, Kristian and Passos, Apolin{\'a}rio},
  booktitle={Proceedings of the IEEE/CVF conference on computer vision and pattern recognition},
  pages={8861--8870},
  year={2024}
}

@article{meng2021sdedit,
  title={Sdedit: Guided image synthesis and editing with stochastic differential equations},
  author={Meng, Chenlin and He, Yutong and Song, Yang and Song, Jiaming and Wu, Jiajun and Zhu, Jun-Yan and Ermon, Stefano},
  journal={arXiv preprint arXiv:2108.01073},
  year={2021}
}

@inproceedings{shalev2026imagerag,
  title={Imagerag: Dynamic image retrieval for reference-guided image generation},
  author={Shalev-Arkushin, Rotem and Gal, Rinon and Bermano, Amit and Fried, Ohad},
  booktitle={International Conference on Learning Representations},
  volume={2026},
  pages={11400--11427},
  year={2026}
}

@inproceedings{zhang2018unreasonable,
  title={The unreasonable effectiveness of deep features as a perceptual metric},
  author={Zhang, Richard and Isola, Phillip and Efros, Alexei A and Shechtman, Eli and Wang, Oliver},
  booktitle={2018 IEEE/CVF conference on computer vision and pattern recognition},
  pages={586--595},
  year={2018},
  organization={IEEE}
}

@article{oquab2023dinov2,
  title={Dinov2: Learning robust visual features without supervision},
  author={Oquab, Maxime and Darcet, Timoth{\'e}e and Moutakanni, Th{\'e}o and Vo, Huy and Szafraniec, Marc and Khalidov, Vasil and Fernandez, Pierre and Haziza, Daniel and Massa, Francisco and El-Nouby, Alaaeldin and others},
  journal={arXiv preprint arXiv:2304.07193},
  year={2023}
}

% %%
% %% If your work has an appendix, this is the place to put it.
\appendix
\clearpage

\newcommand{\grouprow}[1]{%
\multicolumn{11}{@{}l}{\hspace{0.4em}\textcolor{ACMOrange}{\footnotesize\textit{#1}}}%
}

\section{Instantiation of $a_{t,m}$ and $b_{t,m}$ in Eq. \ref{eq:sketch}}

Following DDPM, we parameterize the forward noising process using:
\begin{align}
a_t = \sqrt{\overline{\alpha}_t}, 
\qquad
b_t = \sqrt{1 - \overline{\alpha}_t},
\end{align}
where $\overline{\alpha}_t = \prod_{s=1}^t \alpha_s$ and 
$\beta_t = 1 - \alpha_t$.  
Under the commonly used \texttt{scaled\_linear} beta schedule, as adopted by the PNDM scheduler in Stable Diffusion~1.4, the variance parameter is defined as:
\begin{align}
    \beta_t
    = \Bigl(
        \sqrt{\beta_s}
        + \tfrac{t}{T-1}\bigl(\sqrt{\beta_e} - \sqrt{\beta_s}\bigr)
      \Bigr)^2,
\end{align}
for $t = 0, 1, \dots, T-1$, where $\beta_s$ and $\beta_e$ denote the start and end of the $\beta$ and are set to $0.00085$ and $0.012$ in PNDM scheduler of SD 1.4, respectively.

\paragraph{Affinity-dependent noise modulation.}
To ensure that regions with higher text relevance receive less injected noise, we modulate the terminal noise level $\beta_e$ using the pixel-wise text-visual affinity map $m$. Specifically, we construct a affinity-dependent end-point:
\begin{align}
    \beta_{e,m}
    =
    \beta_s
    + \bigl(1 - f(m)\bigr)\,(\beta_e - \beta_s),
\end{align}
where $m \in [-1,1]^{h\times w}$ is the affinity map and $f(\cdot)$ is a normalization function mapping $m$ into $[0,1]^{h\times w}$ element-wise.

Under this mapping, pixels with larger $m$, which correspond to stronger alignment with the input text, receive a smaller effective $\beta_{e,m}$. As a result, the $\beta_{t,m}$ values across the noising schedule decrease, leading to larger $\alpha_{t,m}=1-\beta_{t,m}$ and an increased cumulative product:
\begin{align}
\overline{\alpha}_{t,m}
= \prod_{s=1}^t \alpha_{s,m}.
\end{align}
Since $a_{t,m} = \sqrt{\overline{\alpha}_{t,m}}$ and $b_{t,m} = \sqrt{1 - \overline{\alpha}_{t,m}}$, a larger $\overline{\alpha}_{T,m}$ leads to a final noisy latent $x_{T,m}$ that preserves more information from the clean latent $x_0$ in relevant regions, enabling heterogeneous noise initialization guided by the affinity map.

\paragraph{Normalization function $f(\cdot)$.}

\begin{figure}[H]
    \centering
    \includegraphics[width=0.8\linewidth]{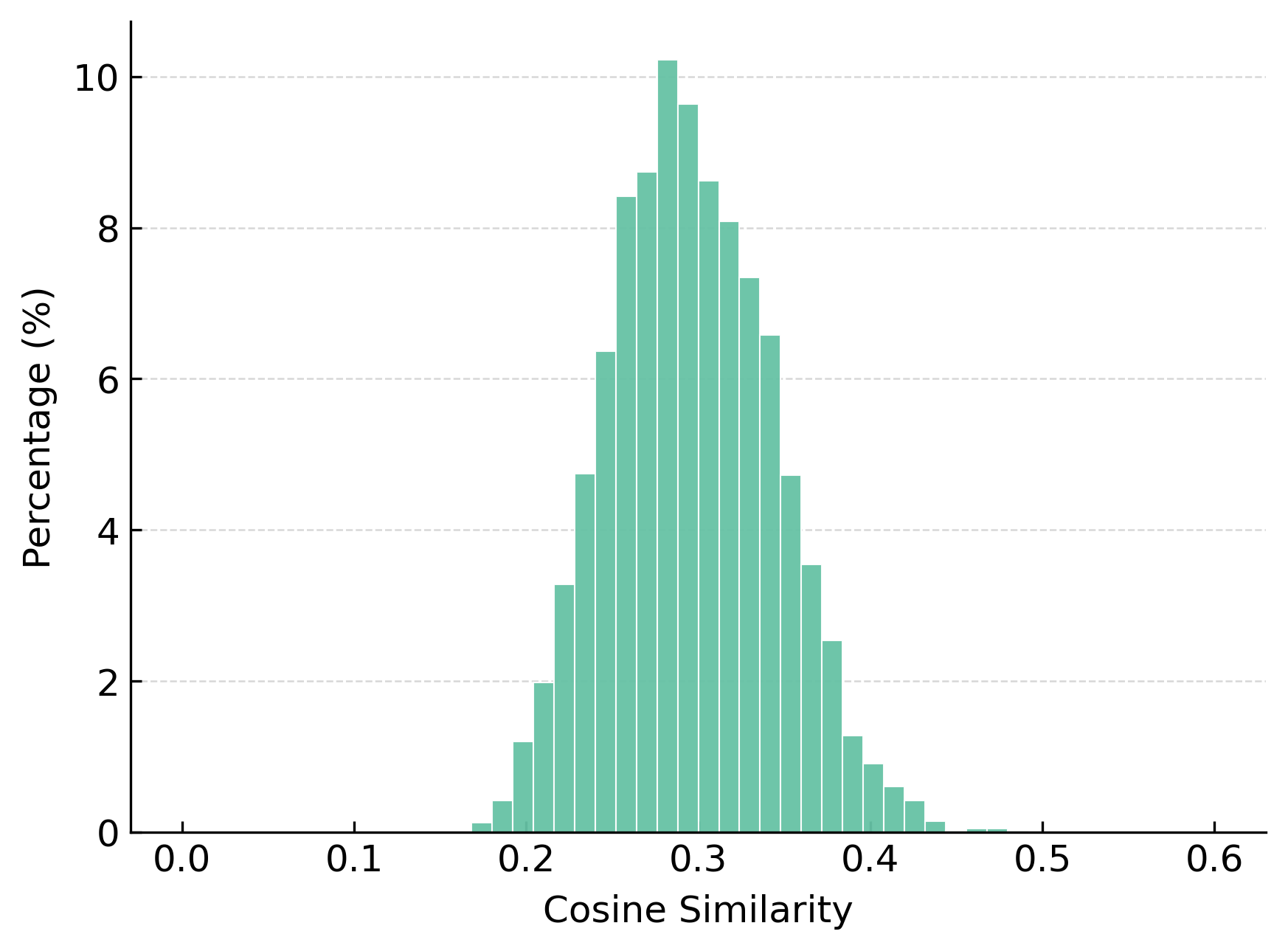}
    \caption{Distribution of the mean CLIP similarity over the top 10\% pixels in each image–text similarity map on a 5k subset of the LAION dataset.}
    \label{fig:clip_dist}
\end{figure}

To modulate the noise schedule on a per-pixel basis, we first require a mapping that transforms the affinity map $m \in [-1,1]^{h\times w}$ into a normalized weight in $[0,1]^{h\times w}$. However, the maximum value of CLIP similarity values used to construct $m$ are typically concentrated near $0.3$, as shown in Fig. \ref{fig:clip_dist}. To enhance the contrast between highly relevant and irrelevant regions and make them uniformly map between 0 and 1, we adopt the following nonlinear normalization function:
\begin{align}
    &f_{\text{raw}}(m)
    = \frac{\arctan\!\bigl(k(m-c)\bigr)}{\arctan(k)} + c, \nonumber\\
    &f(m)
    = \min\!\left(\max\!\left(f_{\text{raw}}(m),\, 0\right),\; 1\right).
\end{align}
where $k$ controls the slope and $c$ determines the activation center of the mapping.
This non-linear mapping expands highly similar regions towards 1 while compressing low similarity scores to 0, with the boundary between expansion and compression at center point $c$.

\begin{figure}[H]
    \centering
    \begin{subfigure}[b]{0.48\linewidth}
        \centering
        \includegraphics[width=\linewidth]{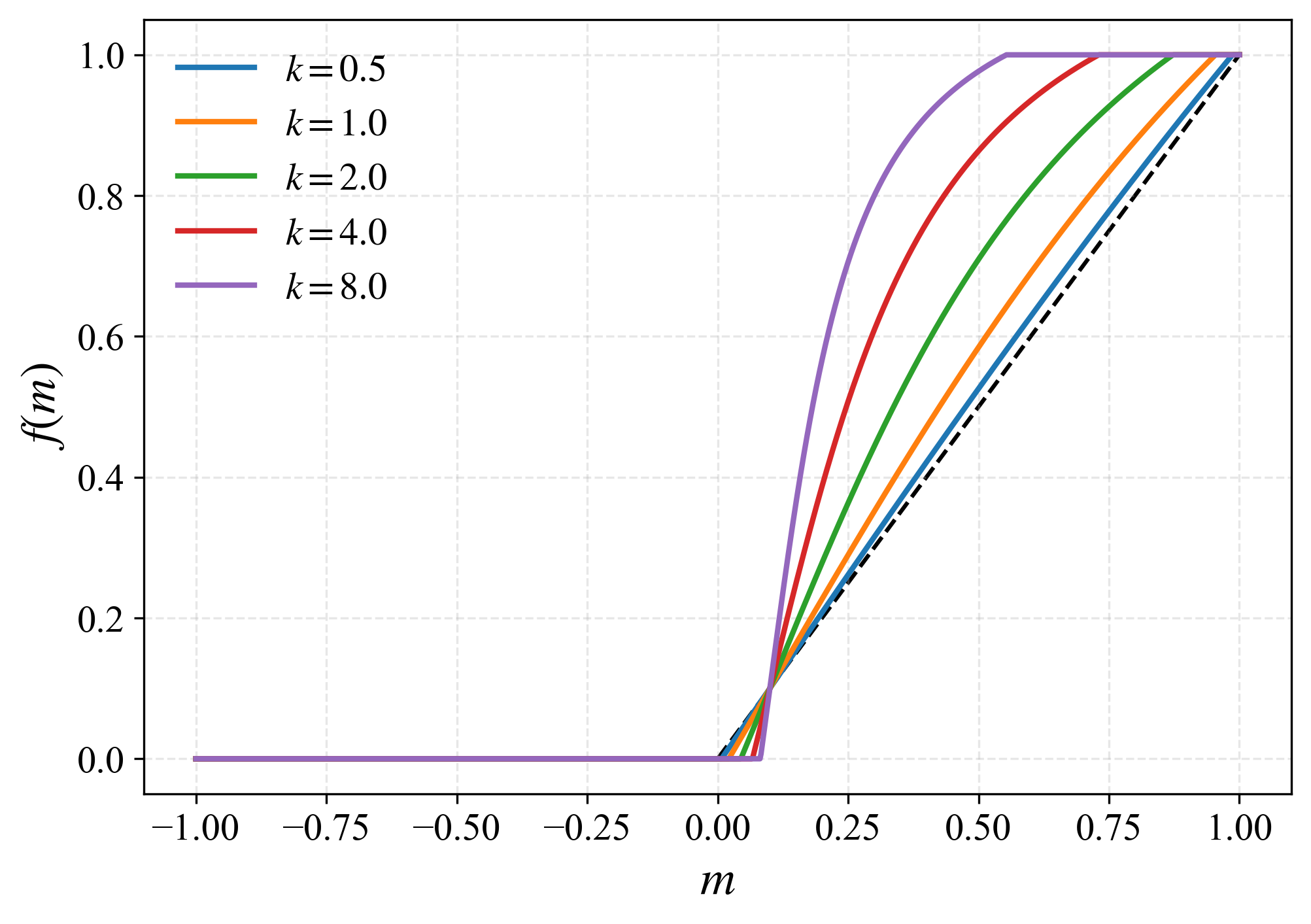}
        \caption{Functions with various slope $k$ when center $c=0.1$.}
    \end{subfigure}
    \hfill
    \begin{subfigure}[b]{0.48\linewidth}
        \centering
        \includegraphics[width=\linewidth]{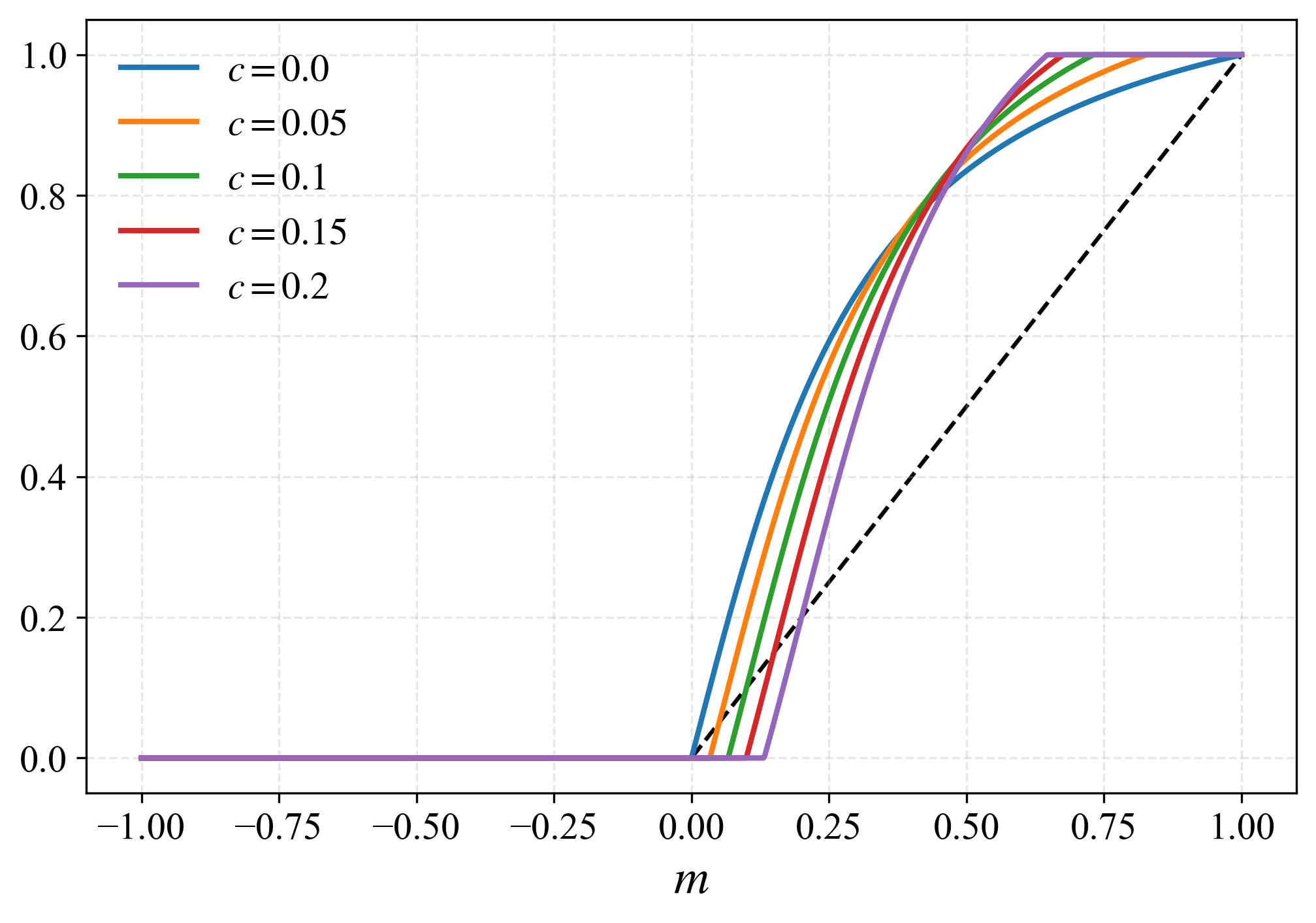}
        \caption{Functions with various center $c$ when slope $k=4$.}
    \end{subfigure}

    \caption{Normalization function $f(\cdot)$ under various slope $k$ and center $c$. The black line indicates that $f(m)=m$.}
    \label{fig:func-kc}
\end{figure}

\begin{figure}[H]
    \centering
    \begin{subfigure}[b]{0.48\linewidth}
        \centering
        \includegraphics[width=\linewidth]{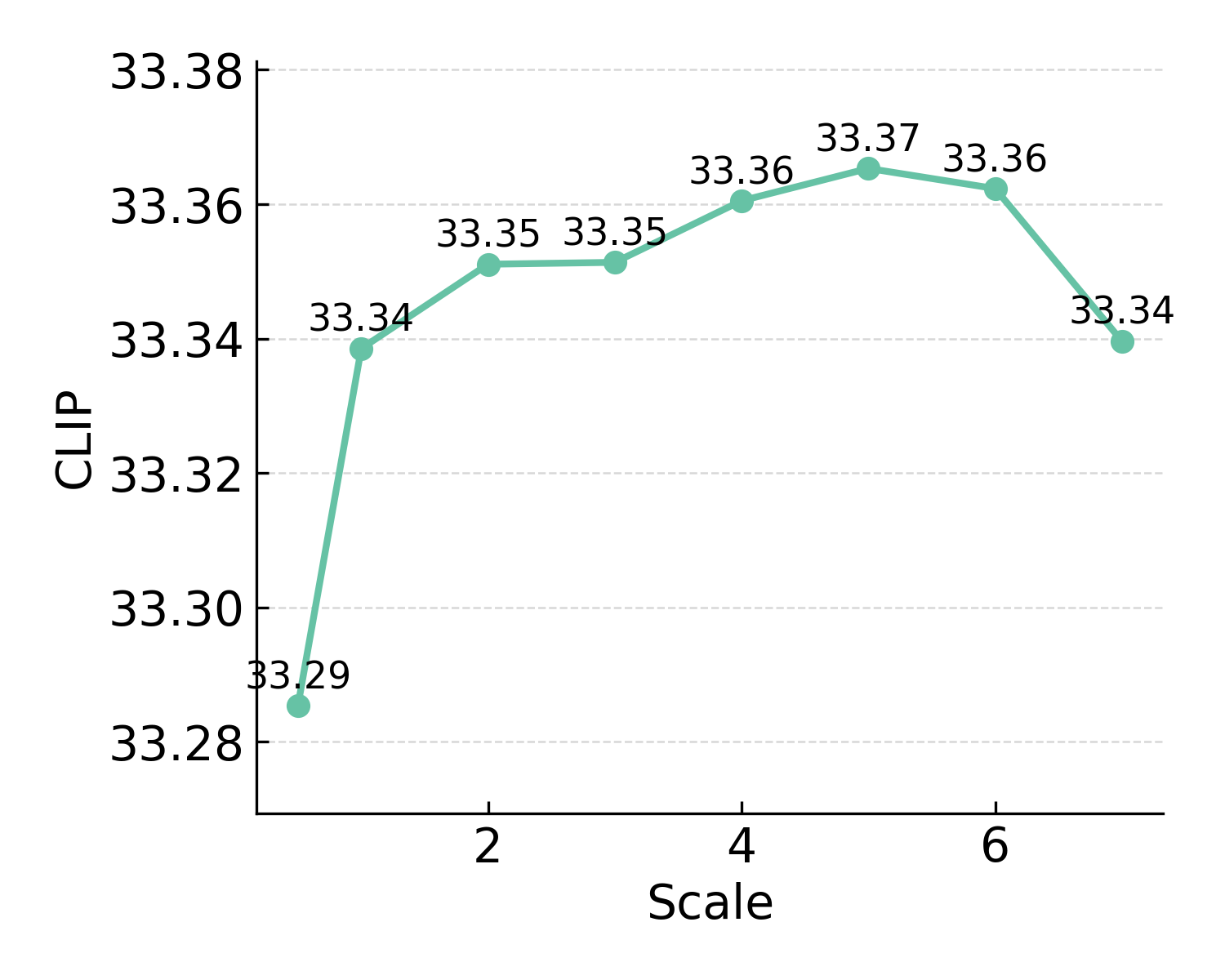}
        \caption{CLIP score under different scale values ($c=0.10$).}
    \end{subfigure}
    \hfill
    \begin{subfigure}[b]{0.48\linewidth}
        \centering
        \includegraphics[width=\linewidth]{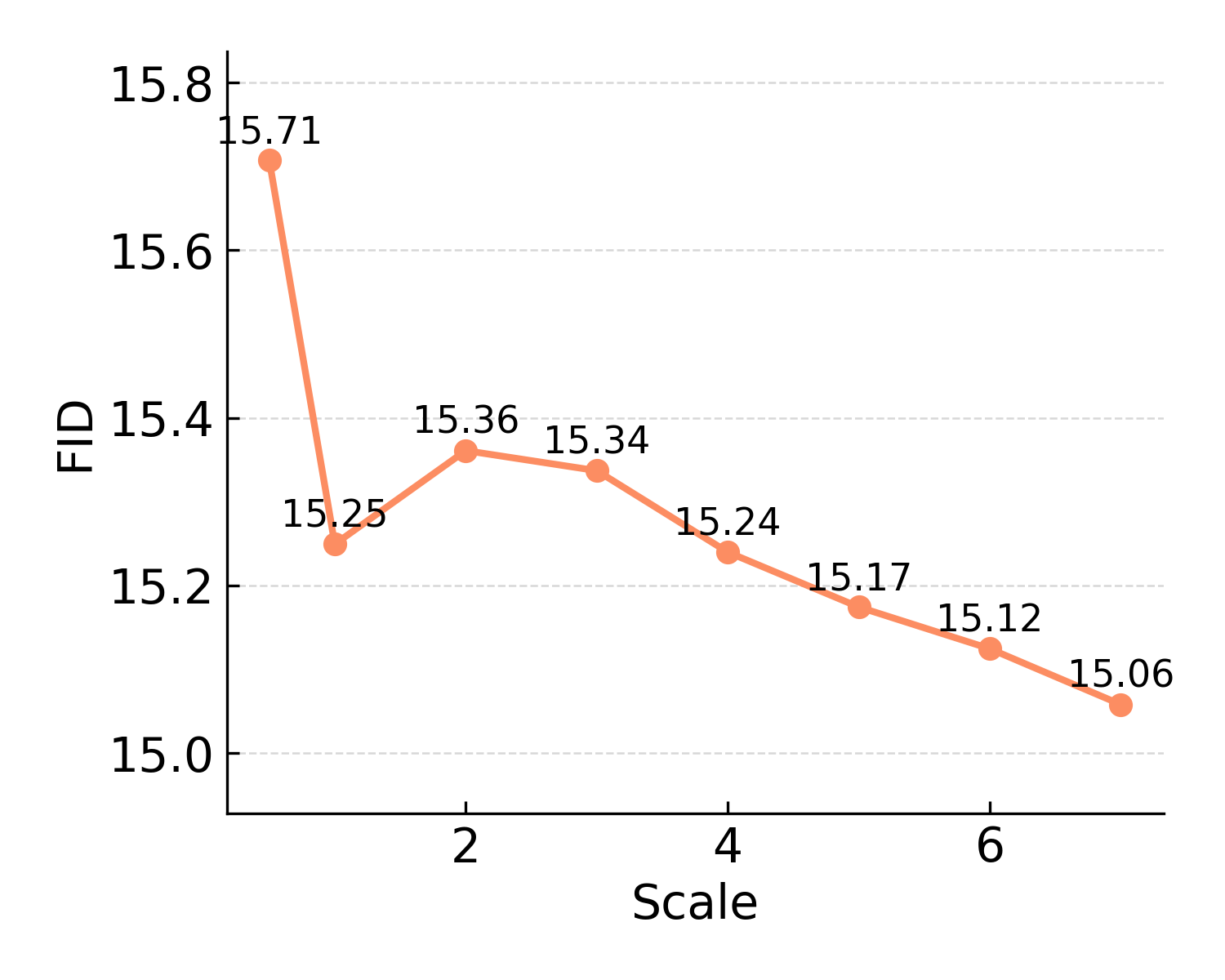}
        \caption{FID under different scale values ($c=0.10$).}
    \end{subfigure}

    \caption{Ablation on the slope parameter $k$ of the normalization function $f(\cdot)$ on a 5k subset of the LAION dataset.}
    \label{fig:ablat-centre0.10}
\end{figure}

\begin{figure}[t!]
    \centering
    \begin{subfigure}[b]{0.48\linewidth}
        \centering
        \includegraphics[width=\linewidth]{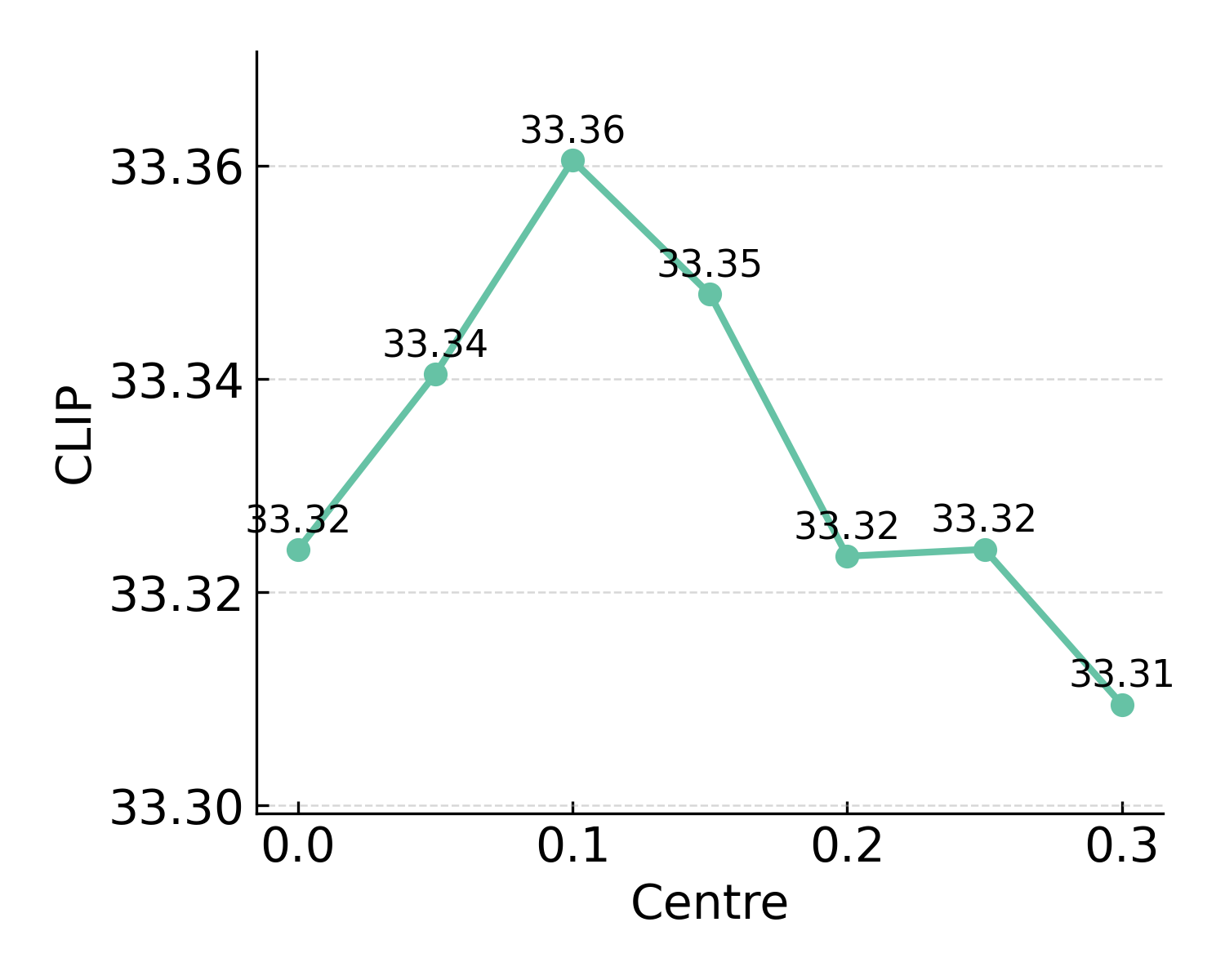}
        \caption{CLIP score under different center values ($k=4.0$).}
    \end{subfigure}
    \hfill
    \begin{subfigure}[b]{0.48\linewidth}
        \centering
        \includegraphics[width=\linewidth]{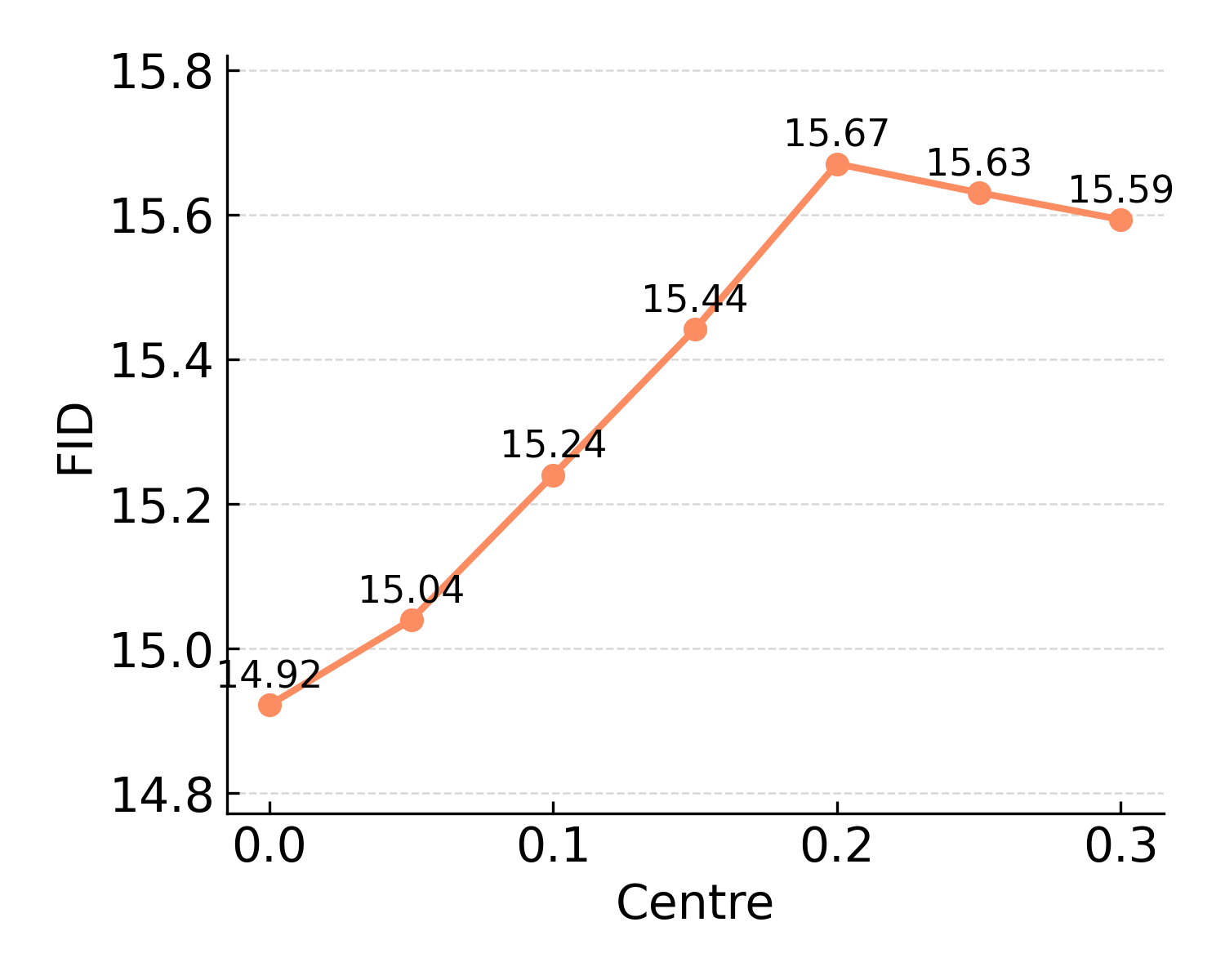}
        \caption{FID under different center values ($k=4.0$).}
    \end{subfigure}

    \caption{Ablation on the center parameter $c$ of the normalization function $f(\cdot)$ on a 5k subset of the LAION dataset.}
    \label{fig:ablat-scale4}
\end{figure}

In our experiments we set $k=4$ and $c=0.1$, which provides a balance between prior strength and generative flexibility. To further understand their roles, we conduct ablation studies on both parameters. The shapes of the corresponding normalization functions are shown in Fig. \ref{fig:func-kc}, and their effects on the generation quality are reported in Fig. \ref{fig:ablat-centre0.10} and Fig. \ref{fig:ablat-scale4}.

As shown in Fig. \ref{fig:func-kc} (a), the slope parameter $k$ primarily controls how strongly the normalization function expands the contrast between different similarity levels. A larger $k$ amplifies high-similarity regions more aggressively, enabling the model to retain stronger visual priors in these areas. As illustrated in Fig. \ref{fig:ablat-centre0.10}, the strengthened priors initially lead to improvements in both CLIP and FID, particularly when $k$ is below 4.0, which shows that moderate reliance on retrieved information enhances both semantic alignment and visual fidelity. However, when $k$ becomes excessively large, the prior dominates the generation process, reducing the model’s creative flexibility. The output then drifts toward the retrieved reference image and away from the input text, leading to a drop in CLIP, while FID continues to decrease because the outputs more closely resemble natural images.

In contrast, the center parameter $c$ determines the boundary between amplified and suppressed regions in the normalization function, as shown in Fig. \ref{fig:func-kc} (b). Fig. \ref{fig:ablat-scale4} demonstrates that a small $c$ mildly suppresses low-similarity regions, effectively filtering out irrelevant structures and improving text alignment, thereby increasing CLIP. However, when $c$ becomes too large, a substantial portion of pixels are classified as low-relevance and overly suppressed. The resulting lack of useful priors makes it difficult for the model to leverage structural cues from retrieved images, causing CLIP to decline.
Meanwhile, FID exhibits a characteristic rise–fall pattern. With mild suppression, the weakened priors cause the generated images to deviate from the natural-image manifold, worsening FID. When $c$ becomes very large, nearly all priors are suppressed and the model effectively falls back to a standard diffusion generator. Its outputs realign with the training distribution, leading to improved FID despite weaker text alignment.

Overall, our choice of the slope $k$ and center $c$ balances the inheritance of retrieved information with the model’s creative space, achieving an effective trade-off between fidelity and flexibility.

\section{Derivation of the coefficients $\kappa_{t,m}$ and $\lambda_{t,m}$ in Eq. \ref{eq:global_constraint}}

As shown in Eq. \ref{eq:noise_pred}, the asynchronous denoising step is
\begin{align}
    x_{t-1}
    = \kappa_{t,m}\, x_t
    + \lambda_{t,m}\, \epsilon_\theta
    + \eta_{t,m}\, \tilde\epsilon_t ,
    \label{eq:eq17}
\end{align}
where $\epsilon_\theta$ is the model-predicted noise at step $t$. For notational brevity, we omit the symbol $\odot$ in subsequent derivations, where all multiplications are implied to be element-wise. For deriving the posterior-matching coefficients, we assume $\epsilon_\theta = \epsilon_t$ inside the training objective.

Using the forward diffusion parameterization $x_t = a_{t,m} x_0 + b_{t,m} \epsilon_t$,
the update Eq. \ref{eq:eq17} becomes
\begin{align}
x_{t-1}
&= \kappa_{t,m}\left(a_{t,m} x_0 + b_{t,m}\epsilon_t\right)
   + \lambda_{t,m}\epsilon_t
   + \eta_{t,m}\tilde\epsilon_t \nonumber\\
&= \kappa_{t,m} a_{t,m} x_0
   + (\kappa_{t,m} b_{t,m} + \lambda_{t,m}) \epsilon_t
   + \eta_{t,m} \tilde\epsilon_t .
\end{align}

To ensure that this transition matches the true diffusion posterior $x_{t-1} = a_{t-1,m} x_0 + b_{t-1,m} \epsilon_t$,
we match both the mean and variance of the two parameterizations.

\paragraph{Mean matching.}
From the $x_0$ term:
\begin{align}
\kappa_{t,m} a_{t,m} = a_{t-1,m}
\quad\Longrightarrow\quad
\boxed{\kappa_{t,m} = \frac{a_{t-1,m}}{a_{t,m}}}.
\end{align}

\paragraph{Variance matching.}
By equating the parameterized variance to the true posterior variance $b_{t-1,m}^2$, we have:
\begin{align}
    (\kappa_{t,m} b_{t,m} + \lambda_{t,m})^2 + \eta_{t,m}^2 &= b_{t-1,m}^2 \nonumber \\
    \implies \kappa_{t,m} b_{t,m} + \lambda_{t,m} &= \sqrt{b_{t-1,m}^2 - \eta_{t,m}^2},
\end{align}
where we select the positive solution to ensure the effective noise coefficient $(\kappa_{t,m} b_{t,m} + \lambda_{t,m})$ remains positive, thereby maintaining a denoising direction consistent with the true posterior.
Therefore, we have:
\begin{align}
\boxed{
\lambda_{t,m}
= -\,\frac{a_{t-1,m} b_{t,m}}{a_{t,m}}
  + \sqrt{\,b_{t-1,m}^2 - \eta_{t,m}^2\,}
}.
\end{align}

\section{Derivation of the Loss Function}

In Section \ref{sec:loss}, we assume that both the true posterior $q(x_{t-1}\mid x_t,x_0)$ and the model distribution $p_\theta(x_{t-1}\mid x_t)$ are diagonal Gaussians:
\begin{align}
    q \sim \mathcal{N}(\mu_q,\Sigma_q),\qquad
    p_\theta \sim \mathcal{N}(\mu_p,\Sigma_p).
\end{align}
Following the asynchronous parameterization introduced in Eq. \ref{eq:sketch} and Eq. \ref{eq:noise_pred}, the transition at step $t$ can be written as:
\begin{align}
    x_{t-1} = u_{t,m}\, x_t + v_{t,m}\, \hat{x}_0 
    + w_{t,m}\, \tilde\epsilon_t,
\end{align}
where the coefficients are:
\begin{align}
    &u_{t,m} = \kappa_{t,m} + \frac{\lambda_{t,m}}{b_{t,m}}, \nonumber \\
    &v_{t,m} = -\frac{\lambda_{t,m} a_{t,m}}{b_{t,m}}, \nonumber \\
    &w_{t,m} = \eta_{t,m},
    \label{eq:uvw}
\end{align}
and $\tilde\epsilon_t\!\sim\!\mathcal N(0,I)$. Hence both posteriors share pixel-wise covariance:
\begin{align}
    \Sigma_q = \Sigma_p = \eta_{t,m}^2 I.
    \label{eq:var_sigma_eta}
\end{align}
The corresponding posterior means are:
\begin{align}
    \mu_q = u_{t,m}\, x_t + v_{t,m}\, x_0,\quad
    \mu_p = u_{t,m}\, x_t + v_{t,m}\, \hat{x}_0.
\end{align}
The KL divergence thus reduces to mean-matching loss:
% Substituting these into the KL divergence gives the pixel-wise mean-matching loss
\begin{align}
    L &= \mathbb{E}_{c,\epsilon,t}\, \mathbb{E}_{m,v}
    \sum_{i} \left[
        \frac{(\mu_p^{(i)} - \mu_q^{(i)})^2}{\Sigma_p^{(i)}}
        + C(t,m)
    \right],
    \label{eq:appendix_loss_start}
\end{align}
where $C(t,m)$ collects entropy terms independent of $\theta$.

By expressing $\hat{x}_0-x_0$ using the DDPM inversion $\hat{x}_0 = \frac{x_t - b_{t,m} \epsilon_\theta}{a_{t,m}}$, we can simplify the mean difference term:
\begin{align}
    &\mu_p^{(i)} - \mu_q^{(i)} \nonumber\\
    =& v_{t,m}^{(i)}\, (\hat{x}_0^{(i)} - x_0^{(i)}) \nonumber\\
    =& v_{t,m}^{(i)}\, \left(\frac{b_{t,m}^{(i)}}{a_{t,m}^{(i)}} \left(\epsilon_t^{(i)} - \epsilon_\theta^{(i)}\right)\right)
\end{align}

\begin{table*}[h]
\centering
\caption{Comparison of PixARG with representative retrieval- and generation-oriented methods under retrieval-augmented diffusion settings using the SD 1.4 backbone on the LAION dataset. DPM++ denotes DPM-Solver++.}
\label{tab:additional_baselines}
% \small
% \vspace{-3mm}
% \renewcommand{\arraystretch}{0.75}
\resizebox{0.8\linewidth}{!}{
\begin{tabular}{lcccccccccc}
\toprule
\multirow{2}{*}{Method} &
\multirow{2}{*}{Steps$\downarrow$} &
\multicolumn{2}{c}{Basic Metrics} &
\multicolumn{2}{c}{Realism \& Diversity} &
\multicolumn{2}{c}{RichHF-18K} &
\multicolumn{3}{c}{Human Preference Metrics} \\
\cmidrule(lr){3-4} \cmidrule(lr){5-6} \cmidrule(lr){7-8} \cmidrule(lr){9-11}
& & CLIP$\uparrow$ & FID$\downarrow$ & Precision$\uparrow$ & Recall$\uparrow$
& Plausibility$\uparrow$ & Overall$\uparrow$ & HPSv2$\uparrow$ & MPS$\uparrow$ & PickScore$\uparrow$ \\
\midrule
SD 1.4 & 50 & 33.10 & 11.49 & 57.07 & 64.73 & 0.6892 & 0.5828 & 24.00 & 9.58 & 19.85 \\
\midrule
\grouprow{\textcolor{ACMRed}{\textbf{Retrieval + Img2Img: generation-only img2img baselines}} using the same retrieved image input as PixARG.} \\
RePaint~\cite{lugmayr2022repaint} & 50 & 30.02 & 10.61 & 73.05 & 45.31 & 0.6200 & 0.5453 & 22.21 & 9.18 & 19.48 \\
LEDITS++~\cite{brack2024ledits++} & 50 & 32.72 & 7.15 & 69.85 & 60.54 & 0.6898 & 0.5846 & 23.70 & 9.73 & 19.93 \\
SDEdit (Tab. 3, Sync.)~\cite{meng2021sdedit} & 15 & 33.30 & 8.64 & 69.31 & 59.49 & 0.6869 & 0.5783 & 23.99 & 9.82 & 19.94 \\
\midrule
\grouprow{\textcolor{ACMRed}{\textbf{Fast Diffusion: generation-only baselines}}, adaptable to diffusion pipelines for acceleration.} \\
LCM~\cite{luo2023latent} & 8 & 29.93 & 37.28 & 49.80 & 21.66 & 0.7912 & 0.6346 & 20.97 & 8.92 & 19.05 \\
PixARG + LCM & 3 & 30.14 & 29.88 & 58.01 & 19.49 & 0.8086 & 0.6361 & 21.09 & 8.95 & 19.24 \\
DPM++~\cite{lu2025dpm} & 15 & 32.94 & 12.96 & 53.64 & 61.29 & 0.6999 & 0.5896 & 23.82 & 9.51 & 19.79 \\
PixARG + DPM++ & 10 & 33.06 & 10.67 & 62.01 & 59.79 & 0.7246 & 0.6039 & 24.43 & 9.93 & 20.05 \\
\midrule
\grouprow{\textcolor{ACMRed}{\textbf{ImageRAG: a retrieval-focused RAG-Diffusion baseline}}, using VLM queries for retrieval and SD+IP-Adapter for generation.} \\
ImageRAG~\cite{shalev2026imagerag} & 50*2+3*VLM & 28.84 & 13.12 & 60.35 & 60.11 & 0.6841 & 0.5838 & 22.10 & 8.69 & 19.08 \\
\midrule
\grouprow{\textcolor{ACMRed}{\textbf{Our default setting: a unified retrieval-generation framework}} integrating retrieved priors into generation.} \\
PixARG (Ours) & 15 & 33.43 & 6.78 & 69.01 & 61.35 & 0.6964 & 0.5860 & 24.40 & 9.96 & 20.06 \\
\bottomrule
\end{tabular}
}
\end{table*}

By substituting these expressions together with Eq. \ref{eq:uvw} and Eq. \ref{eq:var_sigma_eta} into Eq. \ref{eq:appendix_loss_start}, and discarding all terms that do not depend on~$\theta$, we obtain the final objective shown in Eq. \ref{eq:loss_epsilon}.
\begin{align}
    L &= \mathbb{E}_{c,\epsilon,t}\, \mathbb{E}_{m,v} \sum_{i} \left[ \frac{(\mu_p^{(i)} - \mu_q^{(i)})^2}{\Sigma_p^{(i)}} + C(t,m)  \right], \nonumber \\
    &=
        \mathbb{E}_{c,\epsilon,t,m,v}
            \sum_i
                \left[
                    \left(\frac{ v_{t,m}^{(i)}\, b_{t,m}^{(i)} }{ \eta_{t,m}^{(i)}\, a_{t,m}^{(i)} }\right)^2
                    \left(\epsilon_\theta^{(i)} - \epsilon_t^{(i)}\right)^{2}
                    + C(t,m)
                \right]\nonumber\\
    &\propto
    \mathbb{E}_{c,\epsilon,t,m,v}
        \sum_i
            \left(
                \frac{\lambda_{t,m}^{(i)}}{\eta_{t,m}^{(i)}}
            \right)^{2}
            (\epsilon_\theta^{(i)} - \epsilon_t^{(i)})^{2}.
\end{align}

\section{Comparison with related approaches}
Existing retrieval-augmented diffusion methods generally improve either the retrieval process or the generation process, while the interaction between the two stages remains underexplored. In contrast, PixARG explicitly bridges retrieval and generation by deriving a pixel-level affinity map from retrieved references, which subsequently guides pixel-asynchronous initialization and denoising. We further discuss the relationship between PixARG and representative retrieval- or generation-based approaches in Tab.~\ref{tab:additional_baselines}, highlighting that existing approaches typically lack an explicit mechanism to model the spatial correspondence between retrieved images and generation dynamics.

Img2Img methods mainly operate at the generation stage and are designed for tasks requiring strong preservation of input images, such as image editing, repainting, and style transfer. When integrated into a retrieval-augmented generation pipeline, where the retrieved image is provided as the input condition, Img2Img methods naturally encourage the generated image to preserve the retrieved content. While this behavior is beneficial for faithful image transformation, it may limit the flexibility required for open-ended generation. As shown in Tab.~\ref{tab:diversity}, Img2Img-based approaches produce outputs with higher similarity to retrieved images, whereas PixARG better balances retrieved visual priors and generation diversity.

Fast diffusion methods focus on accelerating the generation process by reducing the number of denoising steps. When integrated into a retrieval-augmented generation pipeline, fast diffusion can be combined with Img2Img-style generation, where the retrieved image serves as the input condition. Therefore, fast diffusion is naturally compatible with retrieval augmentation, as the two techniques optimize different aspects of the generation process: fast diffusion reduces the required sampling steps, while retrieved images provide informative visual priors. As shown in Tab.~\ref{tab:additional_baselines}, combining PixARG with accelerated samplers achieves efficient generation while further benefiting from retrieval-guided priors, which improve generation quality beyond the acceleration effect alone.

Within the RAG-diffusion framework, existing methods typically focus on either the retrieval stage or the generation stage. Methods such as ImageRAG mainly improve the retrieval stage by leveraging VLM-based queries to obtain informative visual references, while the retrieval-augmented diffusion methods compared in Tab.~\ref{tab:main_results}~\cite{blattmann2022retrieval,sheynin2022knn,chen2022re} mainly focus on the generation stage by incorporating retrieved images as additional conditioning signals. Different from these approaches, PixARG explicitly models the pixel-level correspondence between retrieved references and the denoising process, enabling adaptive utilization of retrieved information throughout denoising.

\begin{table}[!t]
\centering
\caption{Comparison of retrieval dependency and generation quality between PixARG and Ret.+Img2Img using the SD 1.4 backbone.
LPIPS, DINO, and CLIP-I measure the similarity between generated and retrieved images, where higher LPIPS and lower DINO/CLIP-I indicate reduced retrieval dependency. CLIP-T and FID evaluate generation quality, where higher CLIP-T and lower FID indicate better generation performance.
}
\label{tab:diversity}
\resizebox{0.9\linewidth}{!}{
\begin{tabular}{l|ccc|cc}
\toprule
Method & LPIPS$\uparrow$ & DINO$\downarrow$ & CLIP-I$\downarrow$ & CLIP-T$\uparrow$ & FID$\downarrow$ \\
\midrule
Text2Img & 0.778 & 0.516 & 0.801 & 33.10 & 11.49 \\
Ret.+Img2Img & 0.456 & 0.718 & 0.886 & 33.30 & 8.64 \\
PixARG (Ours) & 0.650 & 0.588 & 0.838 & \textbf{33.43} & \textbf{6.78} \\
\bottomrule
\end{tabular}
}
\end{table}

\begin{table*}[h]
\centering
\caption{Step-matched comparison between PixARG and SD 1.4 on LAION with different denoising steps.}
\label{tab:step_comparison}
\resizebox{0.7\linewidth}{!}{
\begin{tabular}{lcccccccccc}
\toprule
\multirow{2}{*}{Method} &
\multirow{2}{*}{Step} &
\multicolumn{2}{c}{Basic Metrics} &
\multicolumn{2}{c}{Realism \& Diversity} &
\multicolumn{2}{c}{RichHF-18K} &
\multicolumn{3}{c}{Human Preference Metrics} \\
\cmidrule(lr){3-4}
\cmidrule(lr){5-6}
\cmidrule(lr){7-8}
\cmidrule(lr){9-11}
& & CLIP$\uparrow$ & FID$\downarrow$
& Precision$\uparrow$ & Recall$\uparrow$
& Plausibility$\uparrow$ & Overall$\uparrow$
& HPSv2$\uparrow$ & MPS$\uparrow$ & PickScore$\uparrow$ \\
\midrule
SD 1.4 & \multirow{2}{*}{10} & 32.69 & 13.52 & 58.93 & \textbf{57.13} & 0.6660 & 0.5687 & 22.39 & 9.14 & 19.48 \\
PixARG &  & \textbf{33.36} & \textbf{8.40} & \textbf{69.99} & 53.43 & \textbf{0.6886} & \textbf{0.5793} & \textbf{23.97} & \textbf{9.80} & \textbf{19.92} \\
\midrule
SD 1.4 & \multirow{2}{*}{20} & 33.03 & 11.84 & 56.88 & \textbf{62.65} & 0.6796 & 0.5778 & 23.60 & 9.49 & 19.76 \\
PixARG &  & \textbf{33.40} & \textbf{6.83} & \textbf{67.81} & 60.95 & \textbf{0.7057} & \textbf{0.5924} & \textbf{24.59} & \textbf{10.00} & \textbf{20.10} \\
\midrule
SD 1.4 & \multirow{2}{*}{30} & 33.07 & 11.85 & 56.21 & \textbf{64.45} & 0.6817 & 0.5786 & 23.83 & 9.51 & 19.79 \\
PixARG &  & \textbf{33.39} & \textbf{7.00} & \textbf{67.01} & 61.33 & \textbf{0.7077} & \textbf{0.5943} & \textbf{24.77} & \textbf{10.02} & \textbf{20.13} \\
\midrule
SD 1.4 & \multirow{2}{*}{40} & 33.08 & 11.40 & 55.58 & \textbf{64.46} & 0.6880 & 0.5824 & 24.03 & 9.57 & 19.84 \\
PixARG &  & \textbf{33.25} & \textbf{7.36} & \textbf{66.10} & 62.04 & \textbf{0.7126} & \textbf{0.5967} & \textbf{24.63} & \textbf{9.97} & \textbf{20.08} \\
\midrule
SD 1.4 & \multirow{2}{*}{50} & 33.10 & 11.49 & 57.07 & \textbf{64.73} & 0.6892 & 0.5828 & 24.00 & 9.58 & 19.85 \\
PixARG &  & \textbf{33.30} & \textbf{8.02} & \textbf{65.22} & 63.10 & \textbf{0.7165} & \textbf{0.5991} & \textbf{24.76} & \textbf{9.99} & \textbf{20.11} \\
\bottomrule
\end{tabular}
}
\end{table*}

\begin{figure*}[t]
    \centering
    \includegraphics[width=0.7\linewidth]{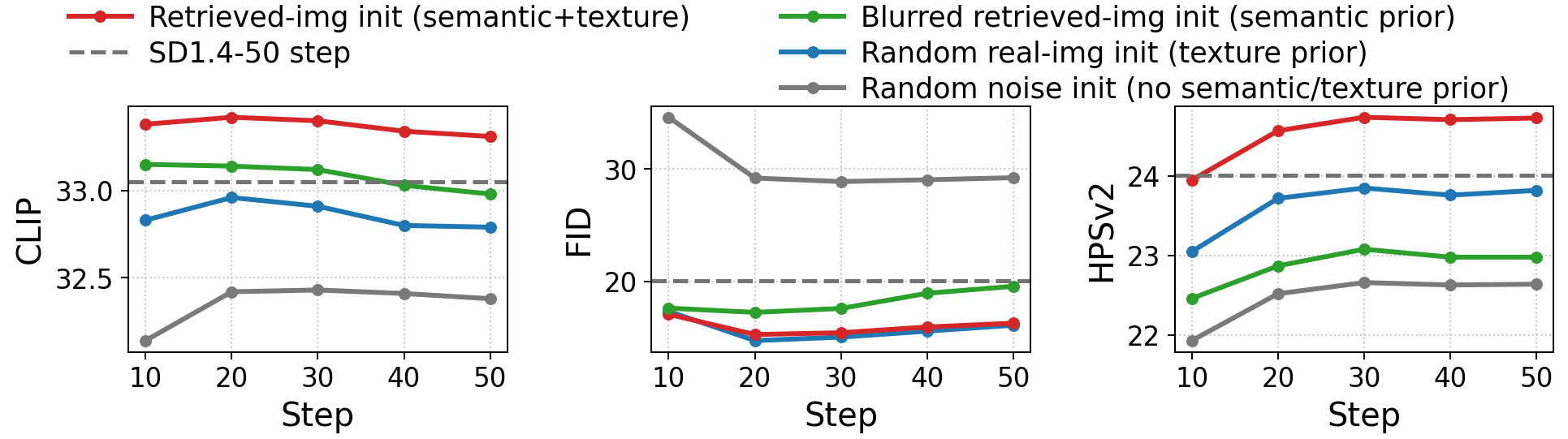}
    \caption{Ablation of retrieval-guided initialization priors on a 5K LAION subset.}
    \label{fig:init_ablation}
\end{figure*}

\section{Diversity analysis}

We analyze how PixARG utilizes retrieved visual priors and compare its retrieval dependency and generation quality with Ret.+Img2Img. Specifically, we evaluate retrieval dependency using perceptual similarity metrics, including LPIPS~\cite{zhang2018unreasonable}, DINO similarity~\cite{oquab2023dinov2}, and CLIP-I~\cite{radford2021learning}, where lower similarity to retrieved images indicates reduced retrieval dependency. Generation quality is measured by CLIP-T~\cite{radford2021learning} and FID~\cite{heusel2017gans}.

As shown in Tab.~\ref{tab:diversity}, compared with Ret.+Img2Img, PixARG generates samples with lower similarity to retrieved images (+0.194 LPIPS, -0.13 DINO, and -0.048 CLIP-I), indicating reduced reliance on retrieved references. Meanwhile, PixARG achieves better text-image alignment and generation quality, with CLIP-T of 33.43 and FID of 6.78. These results demonstrate that PixARG better balances retrieved visual priors and generation flexibility through affinity map-based selective guidance, which adaptively modulates retrieval influence at the pixel level according to relevance weights.

PixARG also improves recall over the SD1.4 T2I baseline by 0.21\% on COYO (Tab.~\ref{tab:main_results_coyo}), 5.95\% on LAION-High (Tab.~\ref{tab:other_models}), and 1.14\% on average across real-user prompts in RedCaps (Tab.~\ref{tab:user_performance}). The consistent recall improvements across diverse retrieval datasets indicate that PixARG preserves generation diversity while benefiting from retrieved visual priors.

\section{Sampling step analysis}

To analyze sampling efficiency under different denoising budgets, we compare PixARG and SD with matched numbers of denoising steps on LAION in Tab.~\ref{tab:step_comparison}. PixARG consistently achieves better performance across different sampling settings. In particular, PixARG requires only around 10--20 denoising steps to match or surpass SD with 50 steps, demonstrating that retrieved visual priors substantially reduce the required denoising iterations. Based on this analysis, we adopt 15 denoising steps as the default sampling setting.

\section{Retrieval-guided speedup analysis}

We analyze how retrieval contributes to efficient generation through initialization ablations in Fig.~\ref{fig:init_ablation}. Retrieved images provide complementary semantic and texture priors through initialization, which reduce the information gap between the initial state and the target distribution, thereby enabling effective generation with fewer denoising steps.

To disentangle the contributions of different retrieval priors, we separately investigate semantic and texture initialization. For semantic priors, blurred retrieved-image initialization suppresses fine-grained textures while preserving high-level structures and global appearance, allowing us to isolate the semantic information provided by retrieval. For texture priors, random real-image initialization preserves natural image statistics without introducing prompt-aligned semantic information, enabling an independent evaluation of texture contribution.

\begin{figure*}[h]
    \centering
    \includegraphics[width=\linewidth]{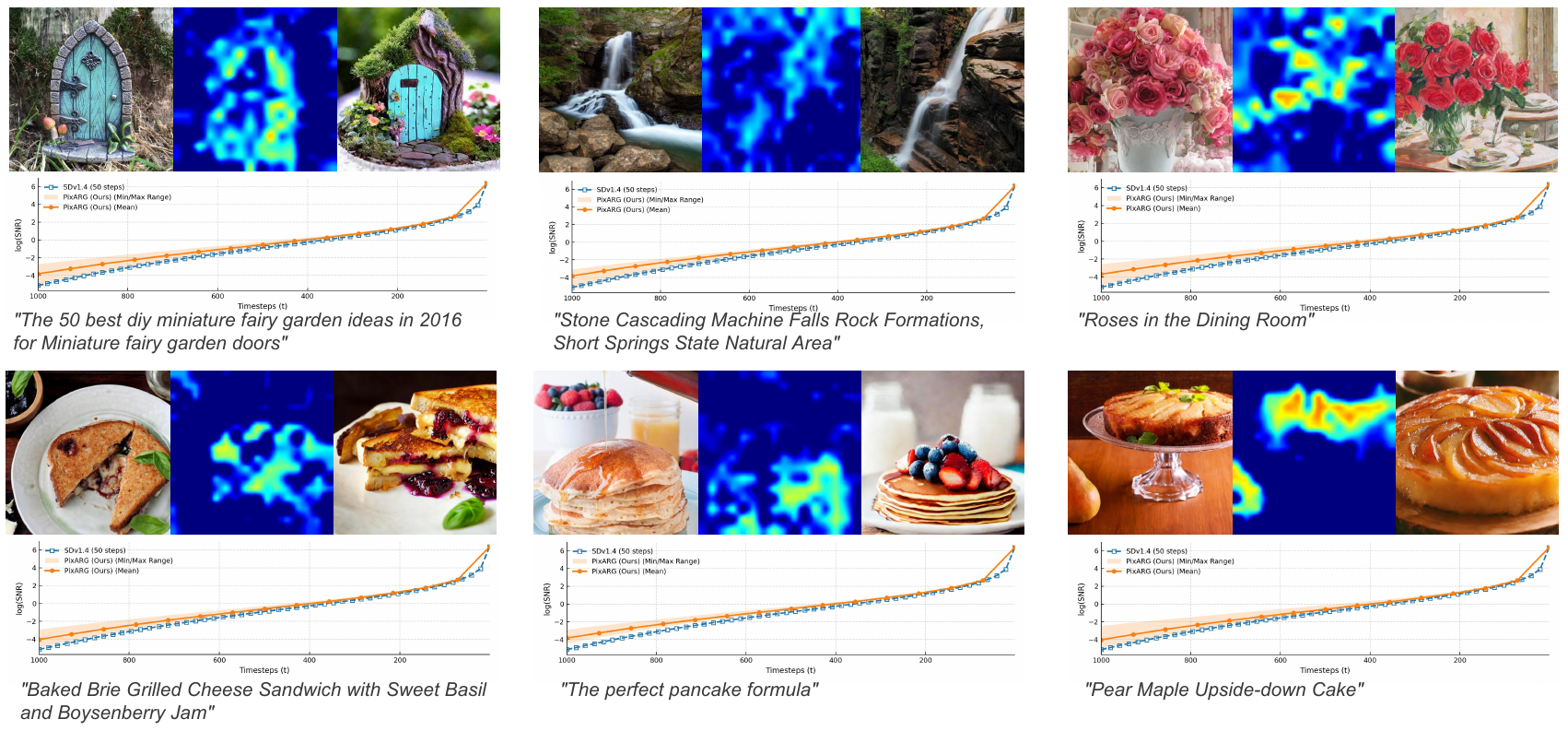}
    \caption{Visualization of PixARG's retrieval and generation results and the corresponding signal-to-noise ratio. The generated results inherit the layout and structural details contained in the affinity-highlighted regions of the retrieved samples while creatively reinterpreting them, maintaining flexibility without losing the underlying structural guidance. Owing to the heterogeneous noise design, each pixel receives its own initial noise level and denoising speed, leading to pixel-wise signal-to-noise ratios that span a certain range.}
    \label{fig:visualization}
\end{figure*}

As shown in Fig.~\ref{fig:init_ablation}, semantic initialization substantially improves CLIP compared with random noise initialization, demonstrating the importance of retrieved semantic structures for maintaining text-image alignment and generation quality. Meanwhile, texture initialization improves HPSv2, indicating that natural image statistics provide additional perceptual benefits. Finally, full retrieved-image initialization, which combines both semantic and texture priors, achieves the best overall performance across CLIP, FID, and HPSv2. These results demonstrate that retrieval provides complementary initialization priors: semantic information guides content generation, while texture information enhances perceptual realism, together reducing the required denoising iterations.

\section{Computational overhead analysis}

To evaluate the efficiency of PixARG, we analyze the computational overhead introduced by retrieval and pixel-level guidance. The reported latency in Tab.~\ref{tab:main_results} is measured end-to-end, including retrieval, retrieved-image encoding, and image generation. PixARG requires 0.367 s/image in total, where retrieval, image encoding, and generation account for 0.3\%, 24.6\%, and 75.1\% of the overall latency, respectively. The retrieval stage introduces negligible overhead, while the majority of the additional computation comes from the one-time encoding of retrieved images to obtain pixel-level guidance.

We further evaluate the memory and storage overhead introduced by PixARG. At batch size 4, PixARG increases RAM and VRAM consumption by 1.38 GB (2.76$\rightarrow$4.14 GB) and 0.95 GB (7.46$\rightarrow$8.41 GB), respectively. The additional memory mainly comes from storing retrieval features and running the retrieval model and VAE image encoder. For retrieval databases, PixARG only requires compact retrieval vectors, with approximately 3 KB storage per image, enabling scalable deployment with large-scale image collections.

Overall, PixARG introduces moderate computational and memory overhead while maintaining efficient end-to-end inference.

\section{Visualization}

Fig.~\ref{fig:visualization} provides a qualitative analysis of our method. 
Unlike conventional diffusion models that follow a single, fixed signal-to-noise ratio (SNR) curve, our approach maintains the SNR within a range. 
This arises from the proposed pixel-wise asynchronous denoising, which allocates the initial noise level and denoising speed adaptively according to the semantic alignment between the query and the retrieved image, ensuring that valuable visual cues are selectively preserved. 
At the initial stage, our logSNR is higher, indicating that generation does not start from pure noise but retains informative visual priors from the retrieved image. 
The generated results inherit both the layout and structural details of the retrieved samples, demonstrating that the visual priors effectively guide the generation process. 
At the same time, the outputs are not direct copies, and the model reinterprets the retrieved content creatively, producing images that remain stylistically consistent yet semantically novel, especially the regions that are discarded due to semantic irrelevance.
These results confirm that our retrieval-augmented diffusion mechanism enables faithful visual grounding while preserving creative flexibility.

\end{document}